\documentclass[APA,Times1COL]{WileyNJDv5} 
\usepackage{xcolor}
\usepackage{caption}
\usepackage{graphicx} % Required for inserting images
\usepackage{tabularx}
\usepackage{booktabs}
\usepackage{multirow}
\usepackage{multicol}
\usepackage{hyperref}
\usepackage{url}
\usepackage{array}
\usepackage{amsmath}

\articletype{Research Article}%

\received{TBD}
\revised{TBD}
\accepted{TBD}
\journal{Epilepsia}
\volume{00}
\copyyear{2026}
\startpage{1}

\begin{document}

\title{Predicting Post-Traumatic Epilepsy from Clinical Records using Large Language Model Embeddings}

\author[uscEE]{Wenhui Cui }
\author[clevcl]{Nicholas Swingle}
\author[uscEE]{Anand A. Joshi}
\author[clevcl]{Dileep Nair} 
\author[uscEE]{Richard M. Leahy}

\authormark{Cui \textsc{et al.}}
\titlemark{Predicting Post-Traumatic Epilepsy from Clinical Records using Large Language Models}

\address[uscEE]{\orgdiv{Ming Hsieh Department of Electrical and Computer Engineering}, \orgname{University of Southern California}, \orgaddress{\state{CA}, \country{USA}}}
\address[clevcl]{\orgdiv{Epilepsy Center}, \orgname{Cleveland Clinic Neurological Institute}, \orgaddress{\state{OH}, \country{USA}}}

\corres{Corresponding author: Richard M. Leahy, Ming Hsieh Department of Electrical and Computer Engineering, 3740 McClintock Avenue Los Angeles, CA 90089-2560, \email{leahy@usc.edu}}

\abstract[Abstract]{Objective: Post-traumatic epilepsy (PTE) is a debilitating neurological disorder that develops after traumatic brain injury (TBI). Early prediction of PTE remains challenging due to heterogeneous clinical data, limited positive cases, and reliance on resource-intensive neuroimaging data. We investigate whether routinely collected acute clinical records alone can support early PTE prediction using language model-based approaches.
Methods: Using a curated subset of the TRACK-TBI cohort, we developed an automated PTE prediction framework that implements pretrained large language models (LLMs) as fixed feature extractors to encode clinical records. Tabular features, LLM-generated embeddings, and hybrid feature representations were evaluated using gradient-boosted tree classifiers under stratified cross-validation.
Results: LLM embeddings achieved performance improvements by capturing contextual clinical information compared to using tabular features alone. The best performance was achieved by a modality-aware feature fusion strategy combining tabular features and LLM embeddings, achieving an AUC-ROC of 0.892 and AUPRC of 0.798.  Acute post-traumatic seizures, injury severity, neurosurgical intervention, and ICU stay are key contributors to the predictive performance.
Significance: These findings demonstrate that routine acute clinical records contain information suitable for early PTE risk prediction using LLM embeddings in conjunction with gradient-boosted tree classifiers. This approach represents a promising complement to imaging-based prediction.
}

\subsection{Key Points}
\begin{itemize}
    \item We describe an LLM-based framework to predict post-traumatic epilepsy using routinely collected acute clinical records without reliance on neuroimaging data.
    \item Acute post-traumatic seizures, injury severity, neurosurgical intervention, and ICU stay were the most relevant data for PTE risk prediction.
    \item Fused feature representations of structured clinical variables and LLM-derived text embeddings achieved the best predictive performance.

\end{itemize}


\maketitle

\section{Introduction}

Post-traumatic epilepsy (PTE) is a debilitating neurological disorder that develops after traumatic brain injury (TBI). As a subtype of acquired epilepsy, PTE is caused by external forces (e.g. accidents or falls) that result in structural or functional brain injury. Clinically, PTE is defined by recurrent and unprovoked seizures occurring more than one week after the initial injury \cite{verellen2010post}. Patients who develop PTE experience reduced functional independence, impaired cognitive performance, and substantially reduced quality of life \cite{burke2021association}. Given these adverse consequences, early identification of patients at high risk for PTE is needed to support timely prevention strategies and therapeutic management.

Prior efforts to predict the risk of developing PTE have approached the problem using scoring algorithms to analyze structured clinical variables and develop a nomogram for PTE risk \cite{wang2021development, wang2021artificial}. Others have used linear logistic regression analysis of clinical data to identify significant predictors of PTE \cite{Raymont2010-ag}. These approaches can predict the risk of developing PTE with impressive accuracy, but the heterogeneity of clinical documentation and inconsistencies between institutions often hinder effective utilization of this rich information \cite{wang2024meditabscalingmedicaltabular}.

Machine learning methods have helped predict the development of a variety of neurological conditions.  Deep learning architectures, including graph neural networks, have been used to classify attention-deficit/hyperactivity disorder (ADHD)  and autism spectrum disorder (ASD) by extracting spatiotemporal features from functional magnetic resonance imaging (fMRI) and identifying abnormalities in functional connectivity \cite{taspinar2024review, ahmed2025summarizing}. In traumatic brain injury, gradient boosting–based models have improved prognostic accuracy by capturing complex nonlinear relationships among clinical variables that are difficult to model using traditional statistical approaches \cite{badjatia2025machine}. In the context of epilepsy, convolutional neural networks and transformer-based models have been applied to electroencephalography (EEG) signals for seizure detection, leveraging spatial feature extraction and temporal sequence modeling to improve performance \cite{zhou2018epileptic, song2022eeg, perez2025artificial}. These approaches have demonstrated the potential to automatically analyze high-dimensional and heterogeneous clinical data and improve predictive performance compared to traditional statistical models. 

Recent advances in large language models (LLMs) offer a promising approach to modeling heterogeneous text-based clinical records \cite{ben2024cpllm, saab2024capabilities, lee2024emergency}. Large language models represent a specialized class of deep learning architectures designed to process sequential data through self attention mechanisms \cite{vaswani2017attention, thirunavukarasu2023large}. Unlike traditional recurrent neural networks (RNNs) that process text sequentially \cite{ghojogh2023recurrentneuralnetworkslong}, LLMs analyze the entire input sequence simultaneously, assigning weights to specific terms to capture long range dependencies and semantic context. By training on  immense amounts of text, LLMs can effectively encode medical knowledge and semantic structure within their parameters to facilitate interpretation of complex clinical narratives without explicit feature engineering. These models can process variable length input and capture contextual relationships within unstructured clinical records, enabling the effective integration of incomplete or inconsistently formatted clinical information \cite{thirunavukarasu2023large, yang2022large}. Using these capabilities, LLMs can facilitate the extraction of clinically meaningful representations from routine medical records to aid the prediction of PTE.

Previous studies applying machine learning to PTE prediction focused primarily on neuroimaging biomarkers \cite{akrami2024prediction, akrami2021prediction, akbar2024advancing, sollee2022artificial}. However, reliance on imaging presents several challenges and may limit early clinical intervention. Imaging data are costly to acquire and often require extensive preprocessing. In multicenter studies, variability in image acquisition protocols requires additional alignment and harmonization procedures. Moreover, image quality can vary substantially between sites. To avoid this potential source of error, we focus our investigation on routinely collected medical records. These records contain diverse and clinically relevant information, including medical history, treatments and medications, post-injury complications, and laboratory test results.

% classical ml models and traditional statistical methods may not be able to handle missing data and inconsistent clinical record categories across cohorts. However, LLMs are agnostic to inconsistent input dimensions and missing data. 

In this work, we investigate the prediction of PTE using acute clinical records from the TRACK-TBI cohort \cite{nelson2021relationship}. Our approach uses a combination of LLM-embeddings to capture the information contained in the clinical record in a succinct manner that can then be used as the basis for PTE-prediction using a low-dimensional classifier, for which purpose we use the gradient-boosted decision tree model, XGBoost \cite{Chen_2016}.   We focus on a well-defined subset of patients with traumatic brain injury and no prior history of seizures or epilepsy with the aim of predicting PTE risk based solely on post-injury clinical information. 

\section{Materials and Methods}
%To capture the multifaceted nature of clinical narratives, text records were organized into demographics and medical history, injury severity, intensive care and neurosurgical interventions, neuroimaging findings, and laboratory test results. Clinical data were represented as short textual summaries to preserve contextual information commonly used in clinical decision-making. Pretrained large language models encoded these clinical narratives into compact representations without any task-specific fine-tuning. This approach allows integration of diverse clinical information while remaining robust to variable documentation practices and missing data. 

\subsection{Dataset}

The Transforming Research and Clinical Knowledge in Traumatic Brain Injury (TRACK-TBI) study \cite{nelson2021relationship} is a large, prospective, multicenter observational cohort designed to standardize and accelerate research in traumatic brain injury. TRACK-TBI enrolled patients presenting with TBI to 18 Level I trauma center emergency departments across the United States between February 2014 and July 2018, with longitudinal follow-up extending to 12 months post-injury. In total, 2,697 participants were enrolled, making TRACK-TBI one of the largest prospective TBI cohorts in the United States. 
Participants were enrolled within 24 hours of injury and were required to have a clinically indicated head CT scan and documentation of TBI consistent with the American Congress of Rehabilitation Medicine definition, including evidence of altered consciousness and/or structural brain injury. Key exclusion criteria included penetrating TBI, non-survivable trauma, severe comorbid medical or neurological conditions, pregnancy, non-English or non-Spanish language proficiency, and enrollment in concurrent interventional trials.

Clinical assessments were conducted acutely and longitudinally by trained and certified examiners. Outcome measures were obtained in person when possible, or by telephone if necessary, at 2 weeks, 3 months, 6 months, and 12 months post-injury, enabling systematic characterization of recovery trajectories. The TRACK-TBI dataset includes detailed demographic information (biological sex, age, race), injury characteristics, and acute clinical measures, such as Glasgow Coma Scale subscales, vital signs, Intensive Care Unit (ICU) admission and length of stay, neurosurgical interventions, and time to surgery. Structured neuroimaging assessments from acute CT scans are available for all participants; these include data on the presence of absence of contusion, hematoma, and hemorrhage. MRI data were collected only in a small subset (around 22\%) of patients. In addition, laboratory results and longitudinal clinical outcomes are recorded, including seizure events and epilepsy diagnoses obtained through followup interviews.

\paragraph{Identification of PTE Cases}
Post-traumatic epilepsy cases were identified using longitudinal interview records within the TRACK-TBI dataset. Subjects were included as PTE cases if they had no documented history of epilepsy prior to TBI and a post-injury epilepsy diagnosis recorded during follow-up. Participants with missing or indeterminate epilepsy diagnosis records were excluded to ensure clear outcome definition.  
Following these criteria, a total of 256 subjects were included in the final cohort, of whom 58 developed PTE and 198 did not develop PTE. Subjects without PTE were designated as the Non-PTE group.

\paragraph{Clinical Feature Selection}
Clinical information was grouped into two categories: structured clinical variables and free-text clinical notes. Structured clinical variables were represented as tabular features, consisting of categorical and numerical values directly extracted from the original clinical records. Feature selection was guided by prior PTE literature \cite{wang2021artificial, wang2021development} to include variables with established or potential relevance to PTE risk. For clinical variables such as Glasgow Coma Scale scores and laboratory test results that were recorded at multiple time points, we compute the first and last recorded values, as well as the minimum, maximum, mean, and standard deviation, to use as tabular features.% However, certain risk factors reported in previous studies \cite{wang2021development}, such as diffuse axonal injury, were not available in our dataset and therefore could not be included in the analysis. 
We also incorporated free-text clinical notes, including neurosurgical operative notes and radiology reports from CT. MRI reports were included when available, though these were typically reported beyond the acute phase and often more than 7 days after injury. For all other clinical variables, only records collected within the first 7 days following injury were used in the analysis.  All structured clinical variables and free-text clinical notes used in this study are summarized in Table~\ref{tab:clinical_features}.

\begin{table}[t]
\centering
\caption{Clinical features used for PTE prediction. Structured clinical variables were extracted from records collected within the first 7 days after injury. Free-text clinical notes were incorporated when available. We group the clinical records into several aspects.}
\label{tab:clinical_features}
\begin{tabular}{p{5cm} p{10cm}}
\hline
\textbf{Clinical Aspect} & \textbf{Variables} \\
\hline

ICU and Surgery &
ICU admission (yes/no), ICU length of stay (days), cranial surgery (yes/no), time to cranial surgery (hours) \\

Early Post-traumatic Seizure &
Seizure within 7 days of injury (yes/no) \\

CT Findings&
Contusion (present/absent/indeterminate), epidural hematoma (present/absent/indeterminate), intracerebral hemorrhage (present/absent/indeterminate), skull fracture (present/absent/indeterminate), subarachnoid hemorrhage (present/absent/indeterminate), Marshall CT score \\

Glasgow Coma Scale &
Glasgow Coma Scale total score; eye, verbal, and motor components \\

Demographics &
Age, sex, race \\

Medical History &
History of epilepsy, prior seizures, neurodegenerative disease, prior neurological illness, transient ischemic attack or stroke, anticoagulant use, antiplatelet use (all yes/no) \\

Laboratory Tests &
Daily creatinine, lactate, hemoglobin, and PaCO\textsubscript{2} \\

\hline
Free-text Clinical Notes &
Cranial surgery operative notes; radiology reports from CT and MRI studies \\
\hline
\end{tabular}
\end{table}

\subsubsection{Tabular Data Serialization}

TRACK-TBI records are summarized within numerous tabular files and exhibit substantial heterogeneity, including inconsistent terminology and missing entries. To address these challenges, we performed systematic data cleaning and consolidation to ensure consistent variable definitions and reliable labeling of PTE. To enable language model-based encoding of structured clinical variables, we converted tabular records into pseudo-clinical notes \cite{lee2024emergency}. Specifically, we grouped variables into clinically meaningful aspects and serialized each aspect into a short natural-language paragraph designed to preserve key clinical content while providing a consistent input format for language models. Each paragraph followed a fixed template. 
We generated a paragraph for each of the following aspects: (1) Glasgow Coma Scale (total and component scores), (2) ICU, surgery performed and operative notes, and acute seizure, (3) CT findings, (4) radiology reports (CT and MRI imaging notes, when available), (5) laboratory tests, (6) medical history and demographics. Missing or unavailable entries were explicitly represented using a standardized missingness token ``NOT\_REPORTED", allowing the model to distinguish absent records consistently. To provide contextual grounding for language model encoding, each aspect paragraph was prefixed with a domain-specific context tag (e.g., Radiology Report, Neurological Examination). These context tags were used to explicitly indicate the clinical source and content of the text, facilitating more consistent interpretation of domain-specific information by the language models. We show examples of serialized clinical aspect paragraphs in Table \ref{tab:serialization_example}.

\begin{table}[t]
\centering
\caption{Example serialized clinical text generated from original TRACK-TBI records. Each paragraph corresponds to one clinical aspect and illustrates the naturalized text input provided to the language models.}
\label{tab:serialization_example}
\begin{tabular}{p{5cm} p{10cm}}
\hline
\textbf{Clinical Aspect} & \textbf{Example Serialized Text} \\
\hline
Glasgow Coma Scale (GCS) &
Neurological Exam (GCS): Worst Total 3-Deep Coma, Best 15. Components (Worst-Best): Eye 1-4, Motor 1-6, Verbal 1-5. \\

\hline
Acute Hospital Course &
Hospital Course: ICU stay 7.2 days. Cranial surgery performed (Decompressive craniectomy). Time to surgery 8.9 hours. Had seizure within 7 days of injury. \\

\hline

CT Findings &
Radiology Report (CT): Findings: Contusion, Epidural Hematoma, Skull Fracture. Absent: Intracerebral Hemorrhage, Subarachnoid Hemorrhage.\\

\hline

Medical History and Demographics &
Patient Demographics: 26-year-old White female. Medical History: No neurological history or anticoagulant/antiplatelet use. \\

\hline
Laboratory Tests &
Laboratory results: Creatinine max value 61.01 occurred 0.02 days after injury, last measurement 0.0, std is 23.70. Hemoglobin max value 7.14 occurred 0.2 days after injury, last measurement 5.15, std is 2.92. Lactate max value 0.0 occurred 0.02 days after injury, last measurement 0.0, std is 0.0. PaCO2 max value 4.79 occurred 0.2 days after injury, last measurement 0.0, std is 1.7.\\
\hline
MRI and CT Imaging Notes &
Radiology Report (Brain): Prior left hemicraniotomy. Again seen multiple foci of susceptibility artifact consistent with diffuse axonal injury. Interval resolution of prior left frontal and temporal and bilateral tentorial extra-axial fluid collections. \\
\hline
\end{tabular}
\end{table}

\subsection{Prediction framework}
\begin{figure}
    \centering
    \includegraphics[width=0.98\linewidth]{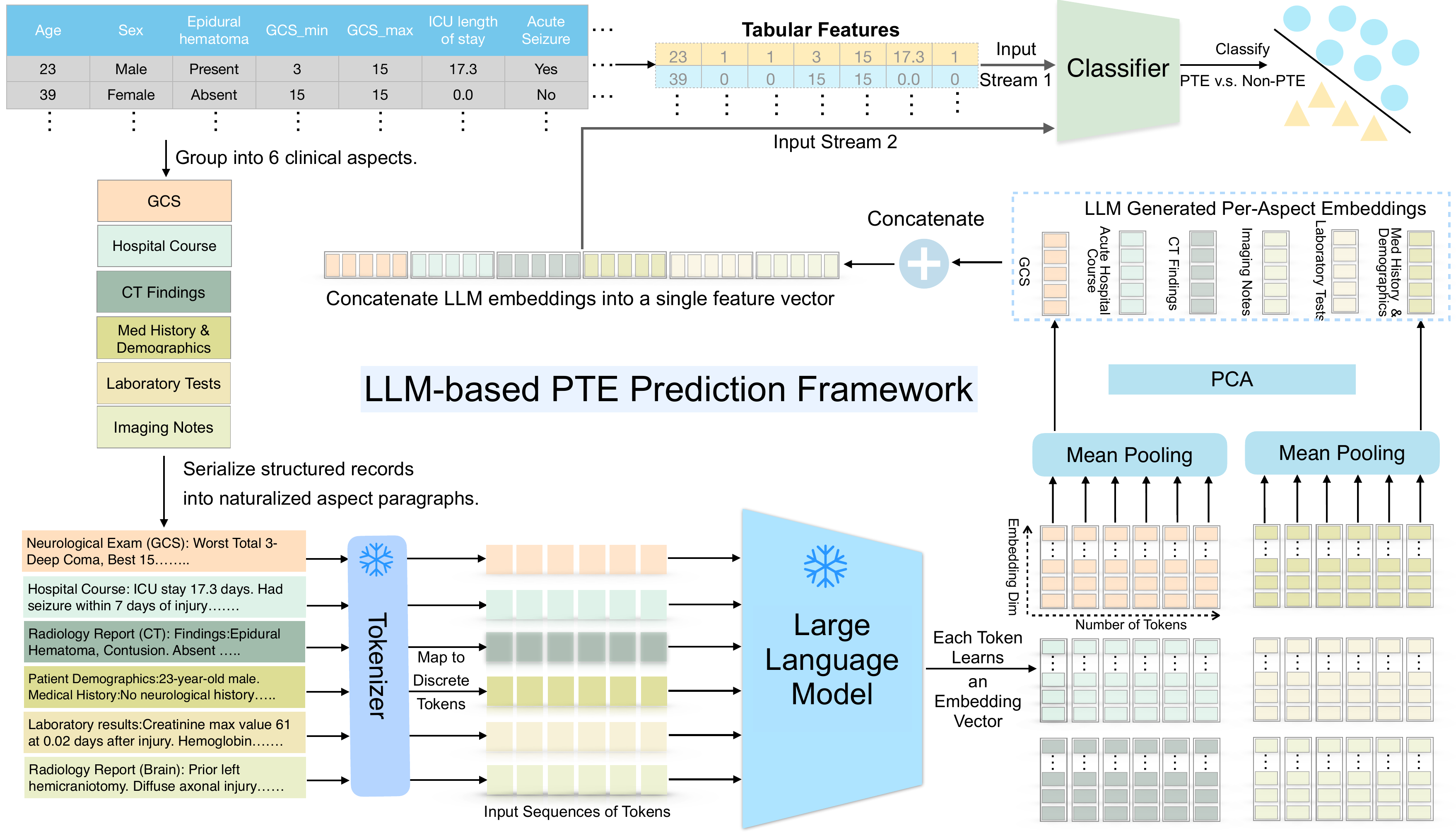}
    \caption{Overview of the proposed framework. Raw acute clinical records are encoded through two parallel input streams. In the first stream, structured clinical variables are represented as tabular features and directly input to the classifier. In the second stream, structured records are grouped into six clinically meaningful aspects, serialized into naturalized paragraphs, tokenized, and encoded by a pretrained LLM to generate per-aspect embeddings. These embeddings are mean-pooled, reduced in dimensionality using PCA, and concatenated into a subject-level feature vector. Each input stream is evaluated independently to predict PTE versus non-PTE. The snowflake symbol indicates that the tokenizer and LLM are frozen (i.e., used as fixed feature extractors without fine-tuning).}
    \label{fig:method}
\end{figure}

We represent clinical information using language model–derived embeddings obtained from serialized clinical records. Pretrained large language models are used as fixed feature extractors to capture contextual information without task-specific fine-tuning. The extracted representations are fed into a lightweight classifier to classify PTE v.s. Non-PTE (see Figure~\ref{fig:method})

\subsubsection{Generation of Text Embeddings from Large Language Models}

To encode clinical text, we apply publicly available pretrained large language models to generate text embeddings from our serialized text inputs. We employed embedding models, corresponding to the transformer encoder part of LLMs, to generate fixed-length numerical representations of clinical text. Each clinical paragraph was first converted into a sequence of tokens using the pretrained tokenizer associated with the corresponding model before being processed by LLMs. Tokenization is the process by which raw text is segmented into smaller units, referred to as tokens, which may represent words, subwords, or punctuation. These tokens serve as the basic input units for transformer-based models, allowing the model to map textual input into numerical representations while preserving linguistic structure. The transformer encoder then maps the input sequence of tokens to contextualized embeddings \cite{vaswani2017attention}. 
% through multi-head self-attention and feed-forward layers 

Both biomedical/clinical–specialized and general-purpose models were tested. 
Representative biomedical/clinical–specialized models included \textbf{BioClinical-ModernBERT} \cite{sounack2025bioclinicalmodernbertstateoftheartlongcontext}, which is based on the BERT \cite{devlin2019bertpretrainingdeepbidirectional} architecture and pretrained using masked language modeling objectives on the largest biomedical and clinical corpus to date gathered from PubMed and PMC, as well as clinical text from 20 datasets. This model incorporated long-context processing and enabled improved representation of medical terminology and clinical context. For general-purpose models, we also integrated several open-source embedding models, including the \textbf{Qwen3} embedding models \cite{qwen3embedding} and \textbf{LLaMA}-based embedding models \cite{babakhin2025llamaembednemotron8buniversaltextembedding} to comprehensively assess the capabilities of textual embeddings. Although the original LLaMA and Qwen models are decoder-only architectures, the embedding models used in this study are specifically designed to operate in an encoder-style mode to generate fixed-length text representations, enabling direct extraction of semantic embeddings rather than autoregressive text generation. General-purpose models are pretrained on broader text corpora to capture general semantic structure. 

Embedding models differ fundamentally from generative LLMs such as ChatGPT or Gemini. Embedding models are encoder-based transformer architectures optimized for representation learning, producing stable and fixed-length vectors that capture semantic and contextual information from text \cite{devlin2019bertpretrainingdeepbidirectional}. In contrast, generative language models are decoder-based or encoder–decoder architectures designed to produce variable length free-text outputs, such as summaries or responses to prompts \cite{radford2019language}. While generative models can be adapted for classification via prompting, their outputs are inherently variable and less suitable for consistent probability prediction.

For each clinical paragraph, the final-layer hidden states produced by the transformer encoder were aggregated using mean pooling to obtain a single fixed-length embedding vector. All LLMs are used as fixed feature extractors without task-specific fine-tuning, which means that all embeddings were generated using frozen models. We designed this strategy for text embedding generation instead of fine-tuning the LLMs to ensure representation stability and reduce the risk of overfitting given the limited size and class imbalance of the PTE dataset. Each input paragraph is encoded into a numerical feature vector that can be directly used as input to a lightweight classifier for downstream prediction tasks. Our proposed approach represents clinical information using separate paragraphs for each clinical aspect, with one embedding vector derived per aspect. This design ensures that information within each clinical aspect is adequately represented and not diluted when aggregating embeddings, particularly when using mean pooling over token-level representations. 

\subsubsection{PTE Classifier}

For each subject, feature representations derived from per-aspect paragraphs are concatenated to form a unified subject-level feature vector. Given the high dimensionality of the combined representations and the relatively small number of subjects, dimensionality reduction was performed using principal component analysis (PCA). PCA was used to project the feature space onto a lower-dimensional subspace while preserving the principal sources of variance, thereby reducing noise and mitigating the risk of overfitting. Based on the number of subjects in the study, the number of retained components was set to 16. To avoid information leakage, PCA was fit exclusively on the training data within each cross-validation fold. 
Classification was performed using a gradient-boosted decision tree model (\textbf{XGBoost}) \cite{Chen_2016}, which was selected for its ability to model nonlinear relationships, handle heterogeneous feature distributions, and remain robust in small-sample settings. To address class imbalance between PTE and non-PTE cases, a weighted loss function was used to assign a higher penalty to misclassification of the minority class (PTE class).

As a baseline comparison, structured clinical variables were provided directly to the classifier as tabular features, with each variable represented as categorical, binary, or numerical value. This baseline was used to assess performance relative to encoding the same clinical information using language model–derived embeddings. In addition to a classifier using exclusively tabular features or language model–based representations, we also include a hybrid approach that integrates both feature types. In the hybrid setting, structured tabular features and text-based embeddings were concatenated to form a unified subject-level representation for classification.

\subsubsection{Evaluation Protocol}

To rigorously evaluate model performance given the class imbalance inherent in our PTE data, we chose clinically meaningful outcome measures. We select Area Under the Receiver Operating Characteristic Curve (AUC-ROC) as the primary metric because it is mathematically invariant to class prevalence. Since the False Positive Rate (calculated solely on negatives) and True Positive Rate (calculated solely on positives) are computed independently, the AUC-ROC provides an unbiased estimate of the model's discriminative ability that is not skewed by the low PTE prevalence of our cohort. We also compute the Area Under the Precision-Recall Curve (AUPRC), which provides a more robust assessment of the model's ability to correctly identify the rare positive class without being biased by the majority of negative cases. To estimate utility in practical settings, we further analyzed precision (Positive Predictive Value, PPV) at fixed recall thresholds of 0.3 and 0.5; these metrics simulate specific clinical operating points, quantifying the model's reliability when configured for high-confidence intervention (targeting the most distinct 30\% of cases) versus a broader screening protocol (capturing 50\% of at-risk patients). These operating points represent clinically meaningful trade-offs: while higher recall prioritizes sensitivity to ensure the maximum number of at-risk patients are identified, the corresponding PPV metric quantifies the efficiency of this screening, reflecting the probability that a flagged individual will genuinely benefit from resource-intensive monitoring or intervention.

Model training and evaluation were conducted using stratified cross-validation to ensure consistent class proportions across folds and to maintain strict separation between training and validation data throughout the modeling pipeline. All metrics were computed within each cross-validation fold, and overall performance is reported as the mean and standard deviation of performance for all folds. 

\subsection{Experiment Design and Setup}\label{Experiment Design and Setup}

We evaluate different experimental setups to assess PTE prediction performance and understand the contribution of individual components. 
\textbf{Baseline.} A permutation study is conducted by randomly shuffling subject labels to confirm that model performance is attributable to clinical signal rather than dataset artifacts or class imbalance. 
\textbf{Feature representation strategies.} Three configurations are evaluated: (1)~\textbf{Tabular Features}: structured clinical variables provided directly to the XGBoost classifier as categorical, binary, or numerical values; (2)~\textbf{LLM Embeds}: text embeddings encoded from serialized clinical records, without any tabular features; and (3) two hybrid approaches. \textbf{Naive Fusion} concatenates all tabular features and LLM embeddings into a single per-subject vector. \textbf{Modality-Aware Fusion} assigns each clinical aspect to the encoding strategy best suited to its data type: CT findings, GCS scores, and laboratory tests are encoded as tabular features, while hospital course (ICU, surgery), imaging notes, medical history, and demographic information are represented as LLM embeddings.

\textbf{Zero-shot prompting.} As an additional comparison, serialized clinical records for all 256 subjects are provided to generative LLMs (GPT-5.2, GPT-5.2 Thinking, Gemini 3 Fast, and Gemini 3 Pro) in a CSV file with each row corresponding to a subject. Models are prompted with: \textit{``You are an expert neurological risk assessment assistant. Task: Predict the probability of patients with traumatic brain injury developing Post-Traumatic Epilepsy (PTE) based on the clinical notes.''} and instructed to output a per-subject PTE probability.

\textbf{Classifier robustness.} In addition to XGBoost, we evaluate the same feature representations with a support vector machine (SVM)~\cite{jakkula2006tutorial} and a multilayer perceptron (MLP)~\cite{popescu2009multilayer} to assess whether LLM embedding gains generalize across classifier families.

\textbf{LLM comparison.} To identify the most effective text encoder, we benchmark biomedical and clinical specialist models (MedBERT, MedCPT-Article-Encoder, PubMedBERT, BioClinical-ModernBERT) and general-purpose embedding models (EmbeddingGemma-300M, Google text-embedding-004, BGE-M3, Qwen-0.6B/4B/8B, LLaMA-8B). We evaluate the PTE prediction performance of embeddings from different LLMs as input to XGBoost classifier. BioClinical-ModernBERT is used as the default encoder for all other experiments.

\textbf{Ablation: input paragraphs and pooling.} We examine the effect of per-aspect versus single-concatenated input paragraphs, and compare three token-level aggregation strategies: mean pooling, $[\mathrm{CLS}]$ token pooling, and max pooling. Given a sequence of input tokens, a transformer encoder produces one embedding vector per token at its final layer. To obtain a single fixed-length representation of the      
  entire input, three aggregation strategies are commonly used. The $[\mathrm{CLS}]$ token approach uses the output of a dedicated classification token prepended to the input, which is trained to summarize the full sequence \cite{devlin2019bertpretrainingdeepbidirectional}. Mean pooling computes the element-wise average across all token embeddings, producing a stable representation that captures information distributed across the sequence \cite{reimers2019sentence}. Max pooling takes the element-wise maximum across token embeddings, retaining the most strongly activated feature in each dimension \cite{conneau2018supervisedlearninguniversalsentence}.

\section{Results}

% \begin{table}[ht]
%     \centering
%     \begin{tabular}{l|c|c|c|c|c}
%         \toprule
%             & AUPRC & AUC-ROC & Precision & Recall (Sensitivity) & F1-score \\
%          \hline
%            GPT 5.2 &  $0.294$ & $0.589$ & $0.417$ & $0.086$ & $0.143$ \\
%         GPT 5.2 - Thinking &  $0.574$ & $0.806$ & $0.732$ & $0.517$ & $0.606$ \\
%         Gemini 3 - Fast & $0.644$ & $0.871$ & $0.632$ & $0.621$ & $0.626$ \\
%            Gemini 3 Pro & $0.510$ & $0.785$ &  $0.488$ & $0.690$ & $0.571$ \\
%     \bottomrule
%     \end{tabular}
%     \caption{Zero-shot Prompting Q\&A LLMs}
%     \label{tab:main_res}
% \end{table}

\subsection{PTE Prediction Results}

Table~\ref{tab:main_res} summarizes the performance of different feature representations and modeling strategies. The permutation test yielded near-random performance across all metrics, confirming that predictive performance elsewhere in the Table is attributable to meaningful clinical information. Using structured tabular features alone resulted in good predictive performance (AUPRC $0.781 \pm 0.086$, AUC-ROC $0.891 \pm 0.045$), highlighting the importance of acute clinical variables for PTE risk estimation. When using the \textbf{BioClinical-ModernBERT }model to generate text embeddings, denoted as LLM Embeds, our approach achieved an improved AUPRC of $0.792 \pm 0.078$, with notably higher PPV at low recall thresholds, indicating improved precision when identifying a subset of patients at higher risk. Directly fusing tabular features with LLM embeddings (Naive Fusion) did not lead to improved performance, but using a modality-aware fusion produced the best overall results as we describe in more detail below. The remainder of Table~\ref{tab:main_res} (rows 6 to 9) are the results of direct prompting of generative-LLMs as described in Section \ref{Experiment Design and Setup}. 

\paragraph{Analysis of Feature Representation Strategy}
Our experiments indicate that different types of clinical information benefit from different representation strategies. Structured variables such as CT findings that are primarily binary indicators of presence, as well as quantitative measures including laboratory test results, and GCS performed best as tabular features. This suggests that while LLMs are powerful semantic extractors, they may struggle to preserve the precise scalar magnitude of numerical values \cite{thawani2021representingnumbersnlpsurvey, golkar2024xval} or be redundant for binary flags when they are embedded using dense vector representations. 

Conversely, complex narrative aspects of the clinical record, particularly ICU course descriptions and cranial surgery notes, provided significantly more discriminability when encoded in the text embedding space. This indicates that the LLM successfully captures contextual dependencies within the combined information. For instance, cranial surgery notes may implicitly contain indicators of severity, such as the presence of an epidural or subdural hematoma. Additionally, demographic information and medical history were found to be represented in a more compact and informative way using LLM embeddings compared to tabular features. Since these variables are primarily encoded as binary indicators, we serialize them into a single natural-language sentence (e.g., no prior medical history), rather than representing the same information with multiple binary flags. This representation enables the model to capture contextual relationships more effectively while reducing feature sparsity.

In our modality-aware fusion strategy, CT findings, GCS, and laboratory tests were encoded using tabular features; hospital course (ICU, surgery), imaging notes, medical history, and demographic information were represented using LLM embeddings. Our modality-aware fusion strategy achieved the best overall performance across all metrics (AUPRC $0.798 \pm 0.073$, AUC-ROC $0.892 \pm 0.042$), with consistent improvements in PPV at all evaluated recall thresholds. A Higher AUC-ROC suggests stronger overall discriminability between PTE and Non-PTE subjects. The increase in AUPRC reflects improved identification of true PTE cases in an imbalanced setting. Higher PPV at predefined recall levels suggests that when the model is used to flag a subset of patients for closer monitoring or follow-up, a greater proportion of those identified are truly at risk of developing PTE. Clinically, this translates to more efficient allocation of surveillance and preventive resources, with fewer low-risk patients subjected to unnecessary interventions while maintaining sensitivity to capture patients at higher risk. 
\begin{table}[ht]
    \centering
     \caption{PTE classification performance across different modeling strategies. Results are reported as mean $\pm$ standard deviation from 5-fold cross-validation. Performance metrics include AUC-ROC, AUPRC, and PPV at fixed recall levels (0.30 and 0.50). LLM-based results in rows 3 to 5 use BioClinical-ModernBERT.}
   \resizebox{\linewidth}{!}{
      
    \begin{tabular}{l|c|c|c|c}
        \toprule
          Method  & AUPRC & AUC-ROC & PPV@Recall$_{0.3}$
 & PPV@Recall$_{0.5}$ \\
         \hline
         1. Permutation Test & $0.311 \pm 0.062$ & $0.532 \pm 0.057$ & $0.389 \pm 0.064$ & $0.262 \pm 0.122$\\
          % Permutation Test & $0.294 \pm 0.057$ & $0.484 \pm 0.069$ & $0.252 \pm 0.078$ & $0.233 \pm 0.041$\\
           % Tabular Only w/o Acute Seizure Info &  $0.738 \pm 0.084$ & $0.830 \pm 0.057$\\
          
        % Tabular Features & $0.785 \pm 0.077$ & $0.891 \pm 0.044$ & $0.629 \pm 0.104$ & $0.686 \pm 0.112$ & $0.643 \pm 0.065$\\ 
         2. Tabular Features & $0.781 \pm 0.086$ & $ 0.891 \pm 0.045$ & $0.927 \pm 0.102$ & $0.879 \pm 0.135$\\ 
         
             3. LLM Embeds &  $ 0.792 \pm 0.078$ & $0.888 \pm 0.047$ & $0.961 \pm 0.065$ & $0.883 \pm 0.129$\\
  
               \hline
             4. Naive Fusion &  $ 0.787 \pm 0.076$ & $0.887 \pm 0.041$ & $0.957 \pm 0.076$ & $0.887 \pm 0.133$ \\
           5. Modality-Aware Fusion & $\mathbf{0.798 \pm 0.075}$ & $\mathbf{0.892 \pm 0.043}$ & $\mathbf{0.963 \pm 0.060}$ & $\mathbf{0.905 \pm 0.122}$\\
           
         % Tabular + Text Embeds  & $0.801 \pm 0.073$ & $0.888 \pm 0.047$ \\
     
        \hline
        Zero-shot Prompting \\
        \hline
         6. GPT 5.2 &  $0.294 \pm 0.098$ & $0.589 \pm 0.077$ & $0.417 \pm 0.155$ & $0.386 \pm 0.180$ \\
        7. GPT 5.2 - Thinking &  $0.622 \pm 0.105$ & $0.808 \pm 0.054$ & $0.732 \pm 0.138$ & $0.692 \pm 0.192$ \\
        8. Gemini 3 - Fast & $0.644 \pm 0.113$ & $0.871 \pm 0.091$ & $0.797 \pm 0.162$ & $0.774 \pm 0.183 $ \\
           9. Gemini 3 Pro & $0.533 \pm 0.087$ & $0.785 \pm 0.056$ &  $0.649 \pm 0.111$ & $0.620 \pm 0.145$  \\
           % DeepSeek-V3.2
    \bottomrule
    \end{tabular}
    }

    \label{tab:main_res}
\end{table}

\paragraph{Zero-Shot Prompting vs. Embeddings}
As shown in Table~\ref{tab:main_res}, zero-shot prompting of generative LLMs resulted in substantially worse performance compared to embedding-based approaches. GPT-5.2 achieved near-random discrimination (AUC-ROC $0.589$), indicating limited effectiveness for this task. While GPT-5.2 with explicit reasoning (“Thinking”) achieved improved performance (AUC-ROC $0.808$), it still exhibited higher variance and consistently lower AUPRC relative to our proposed framework. Similar trends were observed for Gemini models. 
Despite their strong generative and reasoning capabilities, current generative LLMs struggle to perform stable and reliable PTE risk prediction directly from raw clinical records. 
% In contrast, embedding-based representations integrated with conventional machine learning classifiers provide more consistent, reproducible, and clinically meaningful performance for early PTE risk prediction.

\subsubsection{Robustness Across Classifiers}
\begin{figure}[!t]
    \centering
    \includegraphics[width=0.96\linewidth]{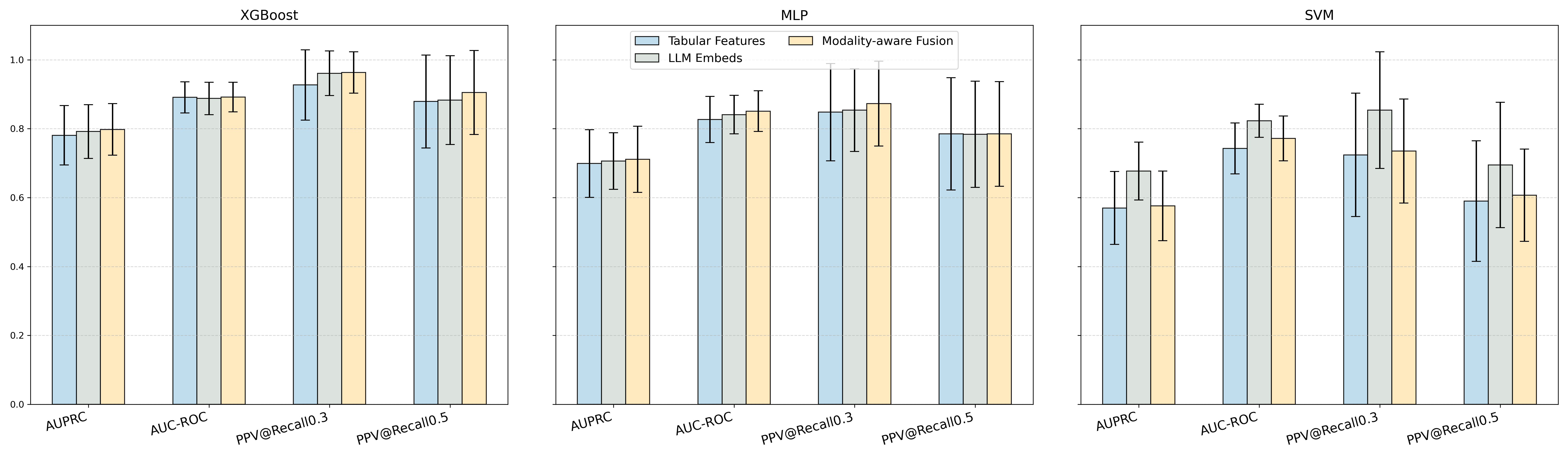}
    \caption{PTE prediction using different classifiers. For SVM and MLP, models incorporating LLM embeddings show consistently improved performance, supporting the robustness of the proposed LLM-based PTE prediction framework. The error bars represent the standard deviations.}
    \label{fig:cls}
\end{figure}

As shown in Figure~\ref{fig:cls}, LLM-generated embeddings consistently improved predictive performance across all classifiers, demonstrating the robustness of LLM-based clinical representations. A Support Vector Machine (SVM) classifier trained on tabular features performed substantially worse than using LLM embeddings. This is likely due to the sensitivity of SVMs to feature scaling and their limited ability to handle non-normalized discrete variables. The tabular features of the TRACK-TBI dataset consist of heterogeneous feature types, including discrete categories, binary indicators, and missing values, which can be challenging for SVMs to handle effectively without extensive preprocessing. In contrast, LLM embeddings provide dense, continuous, and well-normalized representations that embed heterogeneous clinical information into a unified feature space, resulting in the largest relative performance gains for SVM classifiers.  multilayer perceptron (MLP) classifiers also benefited from LLM embeddings, reflecting the advantage of compact semantic representations for gradient-based optimization in small and heterogeneous datasets. XGBoost, by contrast, performed well with tabular features alone due to its native handling of missing values and mixed feature types, leading to smaller relative gains from embeddings. Nevertheless, models incorporating LLM embeddings or fused features consistently achieved the strongest overall performance.

\subsection{Comparison of LLMs and Ablation Study}
Table~\ref{tab:llm_comparison} reports PTE prediction performance across the benchmarked LLMs. BioClinical-ModernBERT achieved the highest performance among medical-specific models (AUPRC $0.792$, AUC-ROC $0.888$), suggesting that domain-specific pretraining is beneficial for understanding clinical context. General-purpose models such as EmbeddingGemma-300M remained highly competitive (AUPRC $0.786$, AUC-ROC $0.890$), indicating that modern, smaller-scale generalist models have sufficient semantic capability to encode clinical narratives effectively. Notably, increasing model size (the number of model parameters) did not strictly correlate with performance. For instance, large-scale models that have proven powerful across a wide range of tasks like Qwen-8B underperformed smaller models, potentially due to overfitting given our limited dataset size. Detailed architectures of the LLMs used in this comparison are described in the Appendix A.1.  

\begin{table}[ht]
    \centering
    \caption{Comparison of PTE prediction performance using text embeddings generated from different pretrained LLMs. Embedding dimensionality is reported for each model. Results are shown as mean $\pm$ standard deviation. Qwen-0.6B/4B/8B correspond to pretrained models with increasing parameter scales of approximately 0.6B, 4B, and 8B parameters.}
    \resizebox{\linewidth}{!}{
    \begin{tabular}{l c c c c c}
        \toprule
        Model & Embedding Dim & AUPRC & AUC-ROC & PPV@Recall$_{0.3}$ & PPV@Recall$_{0.5}$ \\
        \midrule
        \multicolumn{6}{l}{\textit{Biomedical and clinical specialist LLMs}} \\
        \midrule
        MedBERT \cite{9980157} 
            & 768 & $0.769 \pm 0.078$ & $0.877 \pm 0.047$ & $0.953 \pm 0.070$ & $0.857 \pm 0.157$ \\
        MedCPT-Article-Encoder \cite{jin2023medcpt} 
            & 768 & $0.769 \pm 0.084$ & $0.883 \pm 0.047$ & $0.937 \pm 0.094$ & $0.857 \pm 0.146$ \\
        PubMedBERT \cite{mezzetti2023embeddings}
            & 768 & $0.784 \pm 0.079$ & $0.884 \pm 0.050$ & $0.956 \pm 0.078$ & $0.881 \pm 0.138$ \\
        BioClinical-ModernBERT \cite{sounack2025bioclinicalmodernbertstateoftheartlongcontext} 
            & 768 & $\mathbf{0.792 \pm 0.078}$ & $0.888 \pm 0.047$ & $\mathbf{0.961 \pm 0.065}$ & $0.883 \pm 0.129$ \\
        \midrule
        \multicolumn{6}{l}{\textit{General-purpose LLMs}} \\
        \midrule
        EmbeddingGemma-300M \cite{embedding_gemma_2025} 
            & 768 & $0.786 \pm 0.078$ & $\mathbf{0.890 \pm 0.046}$ & $0.952 \pm 0.083$ & $\mathbf{0.891 \pm 0.129}$ \\
        Google text-embedding-004 \cite{google_gemini_embeddings_2024} 
            & 768 & $0.784 \pm 0.073$ & $0.884 \pm 0.044$ & $0.957 \pm 0.073$ & $0.887 \pm 0.131$ \\
        BGE-M3 \cite{bge-m3} 
            & 1024 & $0.766 \pm 0.081$ & $0.878 \pm 0.046$ & $0.949 \pm 0.085$ & $0.856 \pm 0.152$ \\
        Qwen-0.6B \cite{qwen3embedding} 
            & 1024 & $0.772 \pm 0.081$ & $0.877 \pm 0.047$ & $0.950 \pm 0.082$ & $0.867 \pm 0.142$ \\
        Qwen-4B \cite{qwen3embedding} 
            & 2560 & $0.778 \pm 0.084$ & $0.884 \pm 0.047$ & $0.947 \pm 0.087$ & $0.868 \pm 0.156$ \\
        Qwen-8B \cite{qwen3embedding} 
            & 4096 & $0.765 \pm 0.081$ & $0.875 \pm 0.042$ & $0.941 \pm 0.094$ & $0.861 \pm 0.148$ \\
        LLaMA-8B \cite{babakhin2025llamaembednemotron8buniversaltextembedding} 
            & 4096 & $0.774 \pm 0.086$ & $0.882 \pm 0.049$ & $0.945 \pm 0.093$ & $0.865 \pm 0.146$ \\
        \bottomrule
    \end{tabular}
    }
    \label{tab:llm_comparison}
\end{table}

\subsection{Impact of Input Construction and  Pooling Strategies}
Table~\ref{tab:ablation_serialization} summarizes performance under different input construction and aggregation strategies. Per-aspect paragraph encoding with mean pooling consistently outperformed alternatives. Concatenating all six clinical aspects into a single paragraph and encoding them as a single embedding resulted in a substantial performance drop. This finding suggests that collapsing heterogeneous clinical information into a long sequence obscures clinically meaningful distinctions and reduces the model’s ability to capture aspect-specific signals. 
With respect to embedding aggregation strategies, mean pooling consistently outperformed use of the $[\mathrm{CLS}]$ token, which is originally introduced during pretraining to capture a global representation and commonly used in transformer models \cite{dosovitskiy2020image}. Using max pooling also degraded the performance, which selects the maximum value across tokens for each feature dimension. These results indicate that aggregating information across all tokens provides a more stable and informative representation of clinical text, consistent with standard practices in downstream applications of embedding models \cite{reimers2019sentence}.

% Similarly, the template used for serialize the structured text is also important. If we remove the context pref
\begin{table}[ht]
    \centering
    \caption{Impact of per-aspect input paragraphs and pooling strategies on LLM-based PTE prediction. }
    \begin{tabular}{lcc c c}
        \toprule
        Variant & AUPRC & AUC-ROC & PPV@Recall$_{0.3}$ & PPV@Recall$_{0.5}$\\
        \midrule
        Proposed (ours) 
            & $0.792 \pm 0.078$ & $0.888 \pm 0.047$ & $0.961 \pm 0.065$ & $0.883 \pm 0.129$\\
        Concatenated single paragraph 
            & $0.721 \pm 0.091$ & $0.838 \pm 0.060$ & $0.909 \pm 0.129$ & $0.777 \pm 0.158$\\
        % W/o optimized text serialization 
        %     & $0.739 \pm 0.091$ & $0.858 \pm 0.059$ \\
        $[\mathrm{CLS}]$ token pooling 
            & $0.754 \pm 0.088$ & $0.843 \pm 0.064$ & $0.915 \pm 0.113$ & $0.823 \pm 0.139$\\
        Max pooling 
            & $0.772 \pm 0.084$ & $0.866 \pm 0.048$ & $0.932 \pm 0.122$ & $0.849 \pm 0.130$\\
        \bottomrule
    \end{tabular}
    \label{tab:ablation_serialization}
\end{table}

\subsection{Individual Contribution of Clinical Aspects}
To examine the contribution of different clinical information sources, we evaluated PTE prediction performance using each clinical aspect embedding as the only input to the PTE classifier. These embeddings are generated using the BioClinical-ModernBERT model. Table~\ref{tab:aspect_contribution} summarizes the prediction performance for each individual aspect. Among single-aspect models, hospital course information, including ICU admission and duration, neurosurgical intervention, and early post-traumatic seizures, demonstrated the strongest discriminative performance. GCS scores and laboratory test results also showed moderate predictive value, consistent with their established roles as indicators of injury severity. 
In contrast, CT findings alone exhibited limited discriminability, which may reflect suboptimal representation when encoded as text embeddings rather than structured binary indicators. Demographic characteristics and past medical history showed weak standalone predictive performance, which is expected given their indirect and nonspecific relationship to epileptogenesis.
Imaging reports demonstrated poor performance when evaluated across the full cohort, largely due to limited availability. However, restricting the analysis to the subset of subjects with available imaging notes substantially improved performance, suggesting that imaging reports can provide informative signals when consistently available. Overall, the superior performance achieved by combining all clinical aspects underscores the complementary nature of heterogeneous clinical information for PTE prediction.

\begin{table}[ht]
    \centering
    \caption{PTE prediction performance using individual clinical aspects in isolation. Results are reported as mean $\pm$ standard deviation across repeated cross-validation.}
    \begin{tabular}{lcccc}
        \toprule
        Clinical Aspect Used & AUPRC & AUC-ROC & PPV@Recall$_{0.3}$ & PPV@Recall$_{0.5}$ \\
        \midrule
        All aspects (LLM Embeds in Table 3)  
            & $0.792 \pm 0.078$ & $0.888 \pm 0.047$ & $0.961 \pm 0.065$ & $0.883 \pm 0.129$ \\
        Hospital course 
            & $0.737 \pm 0.091$ & $0.872 \pm 0.048$ & $0.894 \pm 0.125$ & $0.824 \pm 0.151$ \\
        Imaging notes (full cohort) 
            & $0.363 \pm 0.061$ & $0.648 \pm 0.061$ & $0.311 \pm 0.059$ & $0.293 \pm 0.031$ \\
        Imaging notes (subset with availability) 
            & $0.554 \pm 0.174$ & $0.750 \pm 0.126$ & $0.760 \pm 0.280$ & $0.587 \pm 0.293$ \\
        Laboratory tests 
            & $0.628 \pm 0.101$ & $0.773 \pm 0.069$ & $0.807 \pm 0.167$ & $0.689 \pm 0.191$ \\
        CT findings 
            & $0.522 \pm 0.101$ & $0.759 \pm 0.068$ & $0.616 \pm 0.161$ & $0.535 \pm 0.142$ \\
       GCS
            & $0.695 \pm 0.091$ & $0.826 \pm 0.063$ & $0.894 \pm 0.129$ & $0.787 \pm 0.169$ \\
   Medical history and Demographics
            & $0.399 \pm 0.077$ & $0.623 \pm 0.071$ & $0.398 \pm 0.138$ & $0.325 \pm 0.081$ \\
        \bottomrule
    \end{tabular}
    \label{tab:aspect_contribution}
\end{table}

\textbf{Contribution of Imaging Notes. }
Imaging report narratives in the TRACK-TBI dataset are typically collected beyond the first week after injury and therefore fall outside the acute period that is the primary focus of this study. Nevertheless, we investigated the contribution of imaging notes when such information is available, to assess their potential value for PTE risk prediction and to provide guidance for future studies. 
In the full cohort, removing imaging notes from the LLM embedding pipeline (denoted as LLM Embeds w/o Imaging Notes in Table~\ref{tab:img_notes}) resulted in comparable performance. This finding is likely attributable to the limited availability of imaging notes, which were present for only 105 of the 256 subjects. To enable a fair comparison, we conducted a secondary analysis restricted to the subset of subjects with available imaging reports (17 PTE cases and 88 non-PTE cases). Within this subset, inclusion of imaging notes led to improved performance in all metrics as shown in Table \ref{tab:img_notes}. 
These results suggest that imaging reports can provide additional discriminative information for identifying patients at risk of PTE when available, although their overall impact is constrained by limited availability and timing outside the acute phase.

\begin{table}[ht]
    \centering
    \caption{Contribution of imaging notes to PTE prediction performance.}
    \begin{tabular}{l|c|c|c|c}
        \toprule
        Method & AUPRC & AUC-ROC & PPV@Recall$_{0.3}$ & PPV@Recall$_{0.5}$ \\
        \midrule
        \multicolumn{5}{l}{\textit{Full cohort (n = 256)}} \\
        \hline
        LLM embeds (with imaging notes) 
            & $0.792 \pm 0.078$ 
            & $0.888 \pm 0.047$ 
            & $0.961 \pm 0.065$ 
            & $0.883 \pm 0.129$ \\
        LLM embeds (without imaging notes) 
            & $0.790 \pm 0.078$ 
            & $0.888 \pm 0.045$ 
            & $0.958 \pm 0.070$ 
            & $0.885 \pm 0.136$ \\
        \midrule
        \multicolumn{5}{l}{\textit{Subset with imaging notes available (n = 105; 17 PTE / 88 non-PTE)}} \\
        \hline
        LLM embeds (w/ imaging notes) 
            & $0.670 \pm 0.175$ 
            & $0.839 \pm 0.104$ 
            & $0.861 \pm 0.215$ 
            & $0.698 \pm 0.282$ \\
        LLM embeds (w/o imaging notes) 
            & $0.628 \pm 0.179$ 
            & $0.829 \pm 0.097$ 
            & $0.823 \pm 0.240$ 
            & $0.644 \pm 0.276$ \\
        \bottomrule
    \end{tabular}
    \label{tab:img_notes}
\end{table}

\subsection{Subgroup Analysis}

% false positives and false negatives, insights. Seems like there is a major overlap of false positives for tabular and text embeds. 
We further evaluated the proposed LLM-based PTE prediction framework across clinically meaningful subgroups categorized by age, ICU admission, neurosurgical intervention, GCS, and acute post-traumatic seizures (within 7 days of injury, see Table~\ref{tab:subgroup_analysis}). Model performance was assessed using AUC-ROC only to quantify discriminative ability within each subgroup, since other metrics are not directly comparable between subgroups.
The proposed framework demonstrated consistently strong discrimination in high-risk subgroups. Patients who underwent cranial surgery and those with severe injury (GCS $<9$) exhibited higher AUC-ROC values. Similarly, the presence of acute post-traumatic seizure was associated with a greater degree or predictive accuracy. Estimates in patients older than 65 years exhibited higher variability, likely due to the limited number of subjects and PTE cases in this subgroup. Overall, the subgroup analysis demonstrates that the proposed LLM-based framework performs most effectively in clinically high-risk populations while maintaining stable performance across diverse patients, supporting its potential applicability for early PTE prediction following traumatic brain injury.
\begin{table}[ht]
\centering
\caption{Subgroup analysis of PTE prediction performance using LLM embeddings as inputs to XGBoost. We aggregated the evaluation results across 5 folds and reported mean $\pm$ standard deviation of AUC-ROC. $\mathrm{N}_{\text{PTE}}$ denotes the number of patients who developed PTE in each subgroup.}
\label{tab:subgroup_analysis}
% \resizebox{\linewidth}{!}{
\begin{tabular}{l|l|c|c|c}
\toprule
Subgroup & Category & N & $\mathrm{N}_{\text{PTE}}$ & AUC-ROC \\
\midrule
Age & $\leq$65 years & 245 & 57 & $0.890 \pm 0.014$ \\
    & $>$65 years    & 11  & 1  & $0.780 \pm 0.261$ \\
\midrule
ICU Admission & No  & 134 & 9  & $0.847 \pm 0.046$ \\
              & Yes & 122 & 49 & $0.851 \pm 0.019$ \\
\midrule
Cranial Surgery Performed & No  & 163 & 19 & $0.801 \pm 0.031$ \\
                & Yes & 93  & 39 & $0.921 \pm 0.020$ \\
\midrule
Minimum GCS Total Score & $<9$   & 78  & 41 & $0.888 \pm 0.025$ \\
            & $\geq$9 & 178 & 17 & $0.797 \pm 0.034$ \\
\midrule
Acute Seizure& No  & 164 & 11 & $0.772 \pm 0.072$ \\
& Yes & 92 & 47 & $0.835 \pm 0.027$ \\
\bottomrule
\end{tabular}
% }
\end{table}

\section{Discussion and Conclusion}

In this study, we introduce an LLM-based framework for predicting post-traumatic epilepsy (PTE) using routinely collected acute clinical records. Evaluated on the TRACK-TBI dataset, our framework demonstrates robust predictive performance, achieving an Area Under the Receiver Operating Characteristic (AUC-ROC) of 0.892 and an Area Under the Precision-Recall Curve (AUPRC) of 0.798. Our results highlight that heterogeneous clinical data are optimally encoded through distinct encoding strategies. Specifically, a modality-aware fusion approach, which integrates structured tabular features with LLM-generated text embeddings yielded the highest overall performance. This confirms that pretrained LLMs can effectively extract clinically relevant semantics from serialized clinical narratives, even without task-specific fine-tuning.

The clinical utility of this framework is demonstrated by its superior performance and its consistency with established PTE literature. At clinically relevant recall thresholds, the model achieved high positive predictive values (PPV@Recall$_{0.3}$ = 0.963; PPV@Recall$_{0.5}$ = 0.905). In a clinical setting, this allows for the reliable identification of high-risk patient subgroups while maintaining a low false-positive rate, which is critical for allocating targeted monitoring and resource-intensive interventions. Furthermore, the model successfully identified established PTE risk factors, including injury severity (indicated by Glasgow Coma Scale and ICU length of stay) and acute seizures within seven days of injury \cite{wang2021artificial, wang2021development}. 

By relying on routine clinical records, our approach addresses significant barriers inherent in previous PTE prediction models. Traditional statistical models often rely strictly on structured clinical variables \cite{wang2021development, wang2021artificial}, making them highly sensitive to missing or inconsistently documented data, such as specific contusion sites. Conversely, recent machine learning efforts have heavily prioritized neuroimaging biomarkers \cite{ayvaz2025predicting, zhou2020machine, la2019machine, akrami2021prediction, cui2023meta, cui2024generalizable}. While imaging provides valuable insights into epileptogenesis, its clinical deployment for acute PTE prediction is constrained by high acquisition costs, inter-site variability in protocols, and complex preprocessing requirements \cite{akrami2024prediction}. By utilizing routine hospital records that are universally available and require minimal harmonization, our framework offers a highly scalable complement to imaging-dependent models.

This work also highlights critical methodological considerations for applying machine learning to rare neurological conditions. Because the availability of PTE data is usually limited, end-to-end fine-tuning of large language models or utilizing high-dimensional neuroimaging features introduces a severe risk of overfitting and poor generalization. We mitigate this issue by employing pretrained LLMs as fixed feature extractors. This strategy successfully leverages the broad linguistic and contextual knowledge embedded within the LLM. Feeding these informative embeddings into a lightweight classifier yields an end-to-end pipeline suited for the limited sample sizes typical of real-world PTE datasets.

Several limitations of this study exist. First, the modest cohort size reflects the inherent rarity of PTE. Track-TBI is a multicenter cohort that has diversity in scanner, injury mechanisms and to some extent reflect heterogeneity of TBI and PTE. However, since TBI and PTE is extremely heterogeneous, further external validation on more independent and larger datasets is desirable Second, the model's performance is tied to the quality and completeness of institutional clinical documentation, which can vary significantly across hospital systems. Finally, because advanced neuroimaging features were unavailable in the TRACK-TBI dataset, we could not evaluate a fully multimodal approach. Future research can investigate the integration of our clinical text embeddings with neuroimaging data to determine if further predictive gains can be achieved when both modalities are available.

In conclusion, language model-based representations of acute clinical records provide an effective and scalable approach for early identification of PTE. By reducing the dependency on advanced imaging and exhaustive manual feature engineering, this framework presents a practical tool for early risk prediction, ultimately facilitating timely patient counseling, targeted follow-up, and proactive management following traumatic brain injury.

\section{Author Contributions}
%let me know if this makes sense for each. https://authorservices.wiley.com/author-resources/Journal-Authors/open-access/credit.html  -  journal wants a CRediT statemnt
\textbf{Wenhui Cui}: conceptualization, data curation, formal analysis, methodology, software, visualization, writing - original draft (lead), writing - review and editing (equal). \textbf{Nicholas Swingle}: conceptualization (supporting), writing - review and editing (equal). \textbf{Anand Joshi:} funding acquisition, data curation, investigation. \textbf{Dileep Nair:} supervision and editing (supporting). \textbf{Richard Leahy: } conceptualization, supervision, resources, writing - review and editing)

\section{Epilepsia Ethical Publication Statement}
We confirm that we have read the Journal’s position on issues involved in ethical publication and affirm that this report is consistent with those guidelines. 

\section{Conflict of Interest Disclosure}
%X has received support from ***
None of the authors has any conflict of interest to disclose.

\section{Acknowledgements}
%do we need to report any funding sources for this work
This work is supported by NIH grant R01 EB026299, and by the DOD grants W81XWH-18-1-061 and HT94252310149. Data used in the preparation of this article reside in the Department of Defense (DOD) and National Institutes of Health (NIH)-supported Federal Interagency Traumatic Brain Injury Research Informatics Systems (FITBIR) in FITBIR-STUDY 0000246. This manuscript reflects the authors' views and does not reflect the opinions or views of the DoD or the NIH.
\bibliography{refs}

\begin{thebibliography}{}

\bibitem [\protect \citeauthoryear {%
Ahmed%
\ \protect \BOthers {.}}{%
Ahmed%
\ \protect \BOthers {.}}{%
{\protect \APACyear {2025}}%
}]{%
ahmed2025summarizing}
\APACinsertmetastar {%
ahmed2025summarizing}%
\begin{APACrefauthors}%
Ahmed, M.%
, Hussain, S.%
, Ali, F.%
, G{\'a}rate-Escamilla, A\BPBI K.%
, Amaya, I.%
, Ochoa-Ruiz, G.%
\BCBL {}\ \BBA {} Ortiz-Bayliss, J\BPBI C.%
\end{APACrefauthors}%
\unskip\
\newblock
\APACrefYearMonthDay{2025}{}{}.
\newblock
{\BBOQ}\APACrefatitle {Summarizing recent developments on autism spectrum disorder detection and classification through machine learning and deep learning techniques} {Summarizing recent developments on autism spectrum disorder detection and classification through machine learning and deep learning techniques}.{\BBCQ}
\newblock
\APACjournalVolNumPages{Applied Sciences}{15}{14}{8056}.
\PrintBackRefs{\CurrentBib}

\bibitem [\protect \citeauthoryear {%
Akbar%
\ \protect \BOthers {.}}{%
Akbar%
\ \protect \BOthers {.}}{%
{\protect \APACyear {2024}}%
}]{%
akbar2024advancing}
\APACinsertmetastar {%
akbar2024advancing}%
\begin{APACrefauthors}%
Akbar, M\BPBI N.%
, Ruf, S\BPBI F.%
, Singh, A.%
, Faghihpirayesh, R.%
, Garner, R.%
, Bennett, A.%
\BDBL {}others%
\end{APACrefauthors}%
\unskip\
\newblock
\APACrefYearMonthDay{2024}{}{}.
\newblock
{\BBOQ}\APACrefatitle {Advancing post-traumatic seizure classification and biomarker identification: Information decomposition based multimodal fusion and explainable machine learning with missing neuroimaging data} {Advancing post-traumatic seizure classification and biomarker identification: Information decomposition based multimodal fusion and explainable machine learning with missing neuroimaging data}.{\BBCQ}
\newblock
\APACjournalVolNumPages{Computerized Medical Imaging and Graphics}{115}{}{102386}.
\PrintBackRefs{\CurrentBib}

\bibitem [\protect \citeauthoryear {%
Akrami%
\ \protect \BOthers {.}}{%
Akrami%
\ \protect \BOthers {.}}{%
{\protect \APACyear {2024}}%
}]{%
akrami2024prediction}
\APACinsertmetastar {%
akrami2024prediction}%
\begin{APACrefauthors}%
Akrami, H.%
, Cui, W.%
, Kim, P\BPBI E.%
, Heck, C\BPBI N.%
, Irimia, A.%
, Jerbi, K.%
\BDBL {}Joshi, A\BPBI A.%
\end{APACrefauthors}%
\unskip\
\newblock
\APACrefYearMonthDay{2024}{}{}.
\newblock
{\BBOQ}\APACrefatitle {Prediction of Post Traumatic Epilepsy Using MR-Based Imaging Markers} {Prediction of post traumatic epilepsy using mr-based imaging markers}.{\BBCQ}
\newblock
\APACjournalVolNumPages{Human Brain Mapping}{45}{17}{e70075}.
\PrintBackRefs{\CurrentBib}

\bibitem [\protect \citeauthoryear {%
Akrami%
, Irimia%
, Cui%
, Joshi%
\BCBL {}\ \BBA {} Leahy%
}{%
Akrami%
\ \protect \BOthers {.}}{%
{\protect \APACyear {2021}}%
}]{%
akrami2021prediction}
\APACinsertmetastar {%
akrami2021prediction}%
\begin{APACrefauthors}%
Akrami, H.%
, Irimia, A.%
, Cui, W.%
, Joshi, A\BPBI A.%
\BCBL {}\ \BBA {} Leahy, R\BPBI M.%
\end{APACrefauthors}%
\unskip\
\newblock
\APACrefYearMonthDay{2021}{}{}.
\newblock
{\BBOQ}\APACrefatitle {Prediction of posttraumatic epilepsy using machine learning} {Prediction of posttraumatic epilepsy using machine learning}.{\BBCQ}
\newblock
\BIn{} \APACrefbtitle {Medical Imaging 2021: Biomedical Applications in Molecular, Structural, and Functional Imaging} {Medical imaging 2021: Biomedical applications in molecular, structural, and functional imaging}\ (\BVOL\ 11600, \BPGS\ 424--430).
\PrintBackRefs{\CurrentBib}

\bibitem [\protect \citeauthoryear {%
Ayvaz%
\ \protect \BOthers {.}}{%
Ayvaz%
\ \protect \BOthers {.}}{%
{\protect \APACyear {2025}}%
}]{%
ayvaz2025predicting}
\APACinsertmetastar {%
ayvaz2025predicting}%
\begin{APACrefauthors}%
Ayvaz, B\BPBI B.%
, Wheelock, J\BPBI R.%
, Jin, D\BPBI S.%
, Appleton, J.%
, Snider, S\BPBI B.%
, Torres-Lopez, V.%
\BDBL {}others%
\end{APACrefauthors}%
\unskip\
\newblock
\APACrefYearMonthDay{2025}{}{}.
\newblock
{\BBOQ}\APACrefatitle {Predicting Post-Traumatic Epilepsy with Automated Contusion Measurements using Acute CT Images: A Competing Risk Approach} {Predicting post-traumatic epilepsy with automated contusion measurements using acute ct images: A competing risk approach}.{\BBCQ}
\newblock
\APACjournalVolNumPages{medRxiv}{}{}{2025--11}.
\PrintBackRefs{\CurrentBib}

\bibitem [\protect \citeauthoryear {%
Babakhin%
\ \protect \BOthers {.}}{%
Babakhin%
\ \protect \BOthers {.}}{%
{\protect \APACyear {2025}}%
}]{%
babakhin2025llamaembednemotron8buniversaltextembedding}
\APACinsertmetastar {%
babakhin2025llamaembednemotron8buniversaltextembedding}%
\begin{APACrefauthors}%
Babakhin, Y.%
, Osmulski, R.%
, Ak, R.%
, Moreira, G.%
, Xu, M.%
, Schifferer, B.%
\BDBL {}Oldridge, E.%
\end{APACrefauthors}%
\unskip\
\newblock
\APACrefYearMonthDay{2025}{}{}.
\newblock
\APACrefbtitle {Llama-Embed-Nemotron-8B: A Universal Text Embedding Model for Multilingual and Cross-Lingual Tasks.} {Llama-embed-nemotron-8b: A universal text embedding model for multilingual and cross-lingual tasks.}
\newblock
\begin{APACrefURL} \url{https://arxiv.org/abs/2511.07025} \end{APACrefURL}
\PrintBackRefs{\CurrentBib}

\bibitem [\protect \citeauthoryear {%
Badjatia%
\ \protect \BOthers {.}}{%
Badjatia%
\ \protect \BOthers {.}}{%
{\protect \APACyear {2025}}%
}]{%
badjatia2025machine}
\APACinsertmetastar {%
badjatia2025machine}%
\begin{APACrefauthors}%
Badjatia, N.%
, Podell, J.%
, Felix, R\BPBI B.%
, Chen, L\BPBI K.%
, Dalton, K.%
, Wang, T\BPBI I.%
\BDBL {}Hu, P.%
\end{APACrefauthors}%
\unskip\
\newblock
\APACrefYearMonthDay{2025}{}{}.
\newblock
{\BBOQ}\APACrefatitle {Machine learning approaches to prognostication in traumatic brain injury} {Machine learning approaches to prognostication in traumatic brain injury}.{\BBCQ}
\newblock
\APACjournalVolNumPages{Current Neurology and Neuroscience Reports}{25}{1}{1--12}.
\PrintBackRefs{\CurrentBib}

\bibitem [\protect \citeauthoryear {%
Ben~Shoham%
\ \BBA {} Rappoport%
}{%
Ben~Shoham%
\ \BBA {} Rappoport%
}{%
{\protect \APACyear {2024}}%
}]{%
ben2024cpllm}
\APACinsertmetastar {%
ben2024cpllm}%
\begin{APACrefauthors}%
Ben~Shoham, O.%
\BCBT {}\ \BBA {} Rappoport, N.%
\end{APACrefauthors}%
\unskip\
\newblock
\APACrefYearMonthDay{2024}{}{}.
\newblock
{\BBOQ}\APACrefatitle {Cpllm: Clinical prediction with large language models} {Cpllm: Clinical prediction with large language models}.{\BBCQ}
\newblock
\APACjournalVolNumPages{PLOS Digital Health}{3}{12}{e0000680}.
\PrintBackRefs{\CurrentBib}

\bibitem [\protect \citeauthoryear {%
Burke%
\ \protect \BOthers {.}}{%
Burke%
\ \protect \BOthers {.}}{%
{\protect \APACyear {2021}}%
}]{%
burke2021association}
\APACinsertmetastar {%
burke2021association}%
\begin{APACrefauthors}%
Burke, J.%
, Gugger, J.%
, Ding, K.%
, Kim, J\BPBI A.%
, Foreman, B.%
, Yue, J\BPBI K.%
\BDBL {}others%
\end{APACrefauthors}%
\unskip\
\newblock
\APACrefYearMonthDay{2021}{}{}.
\newblock
{\BBOQ}\APACrefatitle {Association of Posttraumatic Epilepsy With 1-Year Outcomes After Traumatic Brain Injury} {Association of posttraumatic epilepsy with 1-year outcomes after traumatic brain injury}.{\BBCQ}
\newblock
\APACjournalVolNumPages{JAMA Network Open}{4}{12}{e2140191--e2140191}.
\PrintBackRefs{\CurrentBib}

\bibitem [\protect \citeauthoryear {%
J.~Chen%
\ \protect \BOthers {.}}{%
J.~Chen%
\ \protect \BOthers {.}}{%
{\protect \APACyear {2024}}%
}]{%
bge-m3}
\APACinsertmetastar {%
bge-m3}%
\begin{APACrefauthors}%
Chen, J.%
, Xiao, S.%
, Zhang, P.%
, Luo, K.%
, Lian, D.%
\BCBL {}\ \BBA {} Liu, Z.%
\end{APACrefauthors}%
\unskip\
\newblock
\APACrefYearMonthDay{2024}{}{}.
\newblock
\APACrefbtitle {BGE M3-Embedding: Multi-Lingual, Multi-Functionality, Multi-Granularity Text Embeddings Through Self-Knowledge Distillation.} {Bge m3-embedding: Multi-lingual, multi-functionality, multi-granularity text embeddings through self-knowledge distillation.}
\PrintBackRefs{\CurrentBib}

\bibitem [\protect \citeauthoryear {%
T.~Chen%
\ \BBA {} Guestrin%
}{%
T.~Chen%
\ \BBA {} Guestrin%
}{%
{\protect \APACyear {2016}}%
}]{%
Chen_2016}
\APACinsertmetastar {%
Chen_2016}%
\begin{APACrefauthors}%
Chen, T.%
\BCBT {}\ \BBA {} Guestrin, C.%
\end{APACrefauthors}%
\unskip\
\newblock
\APACrefYearMonthDay{2016}{{\APACmonth{08}}}{}.
\newblock
{\BBOQ}\APACrefatitle {XGBoost: A Scalable Tree Boosting System} {Xgboost: A scalable tree boosting system}.{\BBCQ}
\newblock
\BIn{} \APACrefbtitle {Proceedings of the 22nd ACM SIGKDD International Conference on Knowledge Discovery and Data Mining} {Proceedings of the 22nd acm sigkdd international conference on knowledge discovery and data mining}\ (\BPG~785–794).
\newblock
\APACaddressPublisher{}{ACM}.
\newblock
\begin{APACrefURL} \url{http://dx.doi.org/10.1145/2939672.2939785} \end{APACrefURL}
\newblock
\begin{APACrefDOI} 10.1145/2939672.2939785 \end{APACrefDOI}
\PrintBackRefs{\CurrentBib}

\bibitem [\protect \citeauthoryear {%
Conneau%
, Kiela%
, Schwenk%
, Barrault%
\BCBL {}\ \BBA {} Bordes%
}{%
Conneau%
\ \protect \BOthers {.}}{%
{\protect \APACyear {2018}}%
}]{%
conneau2018supervisedlearninguniversalsentence}
\APACinsertmetastar {%
conneau2018supervisedlearninguniversalsentence}%
\begin{APACrefauthors}%
Conneau, A.%
, Kiela, D.%
, Schwenk, H.%
, Barrault, L.%
\BCBL {}\ \BBA {} Bordes, A.%
\end{APACrefauthors}%
\unskip\
\newblock
\APACrefYearMonthDay{2018}{}{}.
\newblock
\APACrefbtitle {Supervised Learning of Universal Sentence Representations from Natural Language Inference Data.} {Supervised learning of universal sentence representations from natural language inference data.}
\newblock
\begin{APACrefURL} \url{https://arxiv.org/abs/1705.02364} \end{APACrefURL}
\PrintBackRefs{\CurrentBib}

\bibitem [\protect \citeauthoryear {%
Cui%
, Akrami%
, Joshi%
\BCBL {}\ \BBA {} Leahy%
}{%
Cui%
\ \protect \BOthers {.}}{%
{\protect \APACyear {2024}}%
}]{%
cui2024generalizable}
\APACinsertmetastar {%
cui2024generalizable}%
\begin{APACrefauthors}%
Cui, W.%
, Akrami, H.%
, Joshi, A\BPBI A.%
\BCBL {}\ \BBA {} Leahy, R\BPBI M.%
\end{APACrefauthors}%
\unskip\
\newblock
\APACrefYearMonthDay{2024}{}{}.
\newblock
{\BBOQ}\APACrefatitle {Generalizable Representation Learning for fMRI-based Neurological Disorder Identification} {Generalizable representation learning for fmri-based neurological disorder identification}.{\BBCQ}
\newblock
\APACjournalVolNumPages{arXiv preprint arXiv:2412.16197}{}{}{}.
\PrintBackRefs{\CurrentBib}

\bibitem [\protect \citeauthoryear {%
Cui%
, Akrami%
, Zhao%
, Joshi%
\BCBL {}\ \BBA {} Leahy%
}{%
Cui%
\ \protect \BOthers {.}}{%
{\protect \APACyear {2023}}%
}]{%
cui2023meta}
\APACinsertmetastar {%
cui2023meta}%
\begin{APACrefauthors}%
Cui, W.%
, Akrami, H.%
, Zhao, G.%
, Joshi, A\BPBI A.%
\BCBL {}\ \BBA {} Leahy, R\BPBI M.%
\end{APACrefauthors}%
\unskip\
\newblock
\APACrefYearMonthDay{2023}{}{}.
\newblock
{\BBOQ}\APACrefatitle {Meta Transfer of Self-Supervised Knowledge: Foundation Model in Action for Post-Traumatic Epilepsy Prediction} {Meta transfer of self-supervised knowledge: Foundation model in action for post-traumatic epilepsy prediction}.{\BBCQ}
\newblock
\APACjournalVolNumPages{arXiv preprint arXiv:2312.14204}{}{}{}.
\PrintBackRefs{\CurrentBib}

\bibitem [\protect \citeauthoryear {%
Devlin%
, Chang%
, Lee%
\BCBL {}\ \BBA {} Toutanova%
}{%
Devlin%
\ \protect \BOthers {.}}{%
{\protect \APACyear {2019}}%
}]{%
devlin2019bertpretrainingdeepbidirectional}
\APACinsertmetastar {%
devlin2019bertpretrainingdeepbidirectional}%
\begin{APACrefauthors}%
Devlin, J.%
, Chang, M\BHBI W.%
, Lee, K.%
\BCBL {}\ \BBA {} Toutanova, K.%
\end{APACrefauthors}%
\unskip\
\newblock
\APACrefYearMonthDay{2019}{}{}.
\newblock
\APACrefbtitle {BERT: Pre-training of Deep Bidirectional Transformers for Language Understanding.} {Bert: Pre-training of deep bidirectional transformers for language understanding.}
\newblock
\begin{APACrefURL} \url{https://arxiv.org/abs/1810.04805} \end{APACrefURL}
\PrintBackRefs{\CurrentBib}

\bibitem [\protect \citeauthoryear {%
Dosovitskiy%
}{%
Dosovitskiy%
}{%
{\protect \APACyear {2020}}%
}]{%
dosovitskiy2020image}
\APACinsertmetastar {%
dosovitskiy2020image}%
\begin{APACrefauthors}%
Dosovitskiy, A.%
\end{APACrefauthors}%
\unskip\
\newblock
\APACrefYearMonthDay{2020}{}{}.
\newblock
{\BBOQ}\APACrefatitle {An image is worth 16x16 words: Transformers for image recognition at scale} {An image is worth 16x16 words: Transformers for image recognition at scale}.{\BBCQ}
\newblock
\APACjournalVolNumPages{arXiv preprint arXiv:2010.11929}{}{}{}.
\PrintBackRefs{\CurrentBib}

\bibitem [\protect \citeauthoryear {%
Ghojogh%
\ \BBA {} Ghodsi%
}{%
Ghojogh%
\ \BBA {} Ghodsi%
}{%
{\protect \APACyear {2023}}%
}]{%
ghojogh2023recurrentneuralnetworkslong}
\APACinsertmetastar {%
ghojogh2023recurrentneuralnetworkslong}%
\begin{APACrefauthors}%
Ghojogh, B.%
\BCBT {}\ \BBA {} Ghodsi, A.%
\end{APACrefauthors}%
\unskip\
\newblock
\APACrefYearMonthDay{2023}{}{}.
\newblock
\APACrefbtitle {Recurrent Neural Networks and Long Short-Term Memory Networks: Tutorial and Survey.} {Recurrent neural networks and long short-term memory networks: Tutorial and survey.}
\newblock
\begin{APACrefURL} \url{https://arxiv.org/abs/2304.11461} \end{APACrefURL}
\PrintBackRefs{\CurrentBib}

\bibitem [\protect \citeauthoryear {%
Golkar%
\ \protect \BOthers {.}}{%
Golkar%
\ \protect \BOthers {.}}{%
{\protect \APACyear {2024}}%
}]{%
golkar2024xval}
\APACinsertmetastar {%
golkar2024xval}%
\begin{APACrefauthors}%
Golkar, S.%
, Pettee, M.%
, Bietti, A.%
, Eickenberg, M.%
, Cranmer, M.%
, Krawezik, G.%
\BDBL {}Ho, S.%
\end{APACrefauthors}%
\unskip\
\newblock
\APACrefYearMonthDay{2024}{}{}.
\newblock
\APACrefbtitle {xVal: A Continuous Number Encoding for Large Language Models.} {xval: A continuous number encoding for large language models.}
\newblock
\begin{APACrefURL} \url{https://openreview.net/forum?id=OinvjdvPjp} \end{APACrefURL}
\PrintBackRefs{\CurrentBib}

\bibitem [\protect \citeauthoryear {%
Google%
}{%
Google%
}{%
{\protect \APACyear {2024}}%
}]{%
google_gemini_embeddings_2024}
\APACinsertmetastar {%
google_gemini_embeddings_2024}%
\begin{APACrefauthors}%
Google.%
\end{APACrefauthors}%
\unskip\
\newblock
\APACrefYearMonthDay{2024}{}{}.
\newblock
\APACrefbtitle {text-embedding-004.} {text-embedding-004.}
\newblock
\begin{APACrefURL} \url{https://ai.google.dev/gemini-api/docs/embeddings} \end{APACrefURL}
\newblock
\APACrefnote{Accessed: 2025-12-28}
\PrintBackRefs{\CurrentBib}

\bibitem [\protect \citeauthoryear {%
Jakkula%
}{%
Jakkula%
}{%
{\protect \APACyear {2006}}%
}]{%
jakkula2006tutorial}
\APACinsertmetastar {%
jakkula2006tutorial}%
\begin{APACrefauthors}%
Jakkula, V.%
\end{APACrefauthors}%
\unskip\
\newblock
\APACrefYearMonthDay{2006}{}{}.
\newblock
{\BBOQ}\APACrefatitle {Tutorial on support vector machine (svm)} {Tutorial on support vector machine (svm)}.{\BBCQ}
\newblock
\APACjournalVolNumPages{School of EECS, Washington State University}{37}{2.5}{3}.
\PrintBackRefs{\CurrentBib}

\bibitem [\protect \citeauthoryear {%
Jin%
\ \protect \BOthers {.}}{%
Jin%
\ \protect \BOthers {.}}{%
{\protect \APACyear {2023}}%
}]{%
jin2023medcpt}
\APACinsertmetastar {%
jin2023medcpt}%
\begin{APACrefauthors}%
Jin, Q.%
, Kim, W.%
, Chen, Q.%
, Comeau, D\BPBI C.%
, Yeganova, L.%
, Wilbur, W\BPBI J.%
\BCBL {}\ \BBA {} Lu, Z.%
\end{APACrefauthors}%
\unskip\
\newblock
\APACrefYearMonthDay{2023}{}{}.
\newblock
{\BBOQ}\APACrefatitle {MedCPT: Contrastive Pre-trained Transformers with large-scale PubMed search logs for zero-shot biomedical information retrieval} {Medcpt: Contrastive pre-trained transformers with large-scale pubmed search logs for zero-shot biomedical information retrieval}.{\BBCQ}
\newblock
\APACjournalVolNumPages{Bioinformatics}{39}{11}{btad651}.
\PrintBackRefs{\CurrentBib}

\bibitem [\protect \citeauthoryear {%
Kusupati%
\ \protect \BOthers {.}}{%
Kusupati%
\ \protect \BOthers {.}}{%
{\protect \APACyear {2024}}%
}]{%
kusupati2024matryoshkarepresentationlearning}
\APACinsertmetastar {%
kusupati2024matryoshkarepresentationlearning}%
\begin{APACrefauthors}%
Kusupati, A.%
, Bhatt, G.%
, Rege, A.%
, Wallingford, M.%
, Sinha, A.%
, Ramanujan, V.%
\BDBL {}Farhadi, A.%
\end{APACrefauthors}%
\unskip\
\newblock
\APACrefYearMonthDay{2024}{}{}.
\newblock
\APACrefbtitle {Matryoshka Representation Learning.} {Matryoshka representation learning.}
\newblock
\begin{APACrefURL} \url{https://arxiv.org/abs/2205.13147} \end{APACrefURL}
\PrintBackRefs{\CurrentBib}

\bibitem [\protect \citeauthoryear {%
Lee%
\ \protect \BOthers {.}}{%
Lee%
\ \protect \BOthers {.}}{%
{\protect \APACyear {2024}}%
}]{%
lee2024emergency}
\APACinsertmetastar {%
lee2024emergency}%
\begin{APACrefauthors}%
Lee, S\BPBI A.%
, Jain, S.%
, Chen, A.%
, Ono, K.%
, Fang, J.%
, Rudas, A.%
\BCBL {}\ \BBA {} Chiang, J\BPBI N.%
\end{APACrefauthors}%
\unskip\
\newblock
\APACrefYearMonthDay{2024}{}{}.
\newblock
{\BBOQ}\APACrefatitle {Emergency department decision support using clinical pseudo-notes} {Emergency department decision support using clinical pseudo-notes}.{\BBCQ}
\newblock
\APACjournalVolNumPages{arXiv preprint arXiv:2402.00160}{}{}{}.
\PrintBackRefs{\CurrentBib}

\bibitem [\protect \citeauthoryear {%
Mezzetti%
}{%
Mezzetti%
}{%
{\protect \APACyear {2023}}%
}]{%
mezzetti2023embeddings}
\APACinsertmetastar {%
mezzetti2023embeddings}%
\begin{APACrefauthors}%
Mezzetti, D.%
\end{APACrefauthors}%
\unskip\
\newblock
\APACrefYearMonthDay{2023}{Oct}{18}.
\newblock
\APACrefbtitle {Embeddings for Medical Literature: A Strong Baseline Model for Semantic Search and More.} {Embeddings for medical literature: A strong baseline model for semantic search and more.}
\newblock
\begin{APACrefURL} \url{https://medium.com/neuml/embeddings-for-medical-literature-74dae6abf5e0} \end{APACrefURL}
\newblock
\APACrefnote{Medium}
\PrintBackRefs{\CurrentBib}

\bibitem [\protect \citeauthoryear {%
Nelson%
\ \protect \BOthers {.}}{%
Nelson%
\ \protect \BOthers {.}}{%
{\protect \APACyear {2021}}%
}]{%
nelson2021relationship}
\APACinsertmetastar {%
nelson2021relationship}%
\begin{APACrefauthors}%
Nelson, L\BPBI D.%
, Kramer, M\BPBI D.%
, Joyner, K\BPBI J.%
, Patrick, C\BPBI J.%
, Stein, M\BPBI B.%
, Temkin, N.%
\BDBL {}others%
\end{APACrefauthors}%
\unskip\
\newblock
\APACrefYearMonthDay{2021}{}{}.
\newblock
{\BBOQ}\APACrefatitle {Relationship between transdiagnostic dimensions of psychopathology and traumatic brain injury (TBI): A TRACK-TBI study.} {Relationship between transdiagnostic dimensions of psychopathology and traumatic brain injury (tbi): A track-tbi study.}{\BBCQ}
\newblock
\APACjournalVolNumPages{Journal of abnormal psychology}{130}{5}{423}.
\PrintBackRefs{\CurrentBib}

\bibitem [\protect \citeauthoryear {%
Perez-Sanchez%
\ \protect \BOthers {.}}{%
Perez-Sanchez%
\ \protect \BOthers {.}}{%
{\protect \APACyear {2025}}%
}]{%
perez2025artificial}
\APACinsertmetastar {%
perez2025artificial}%
\begin{APACrefauthors}%
Perez-Sanchez, A\BPBI V.%
, Valtierra-Rodriguez, M.%
, De-Santiago-Perez, J\BPBI J.%
, Perez-Ramirez, C\BPBI A.%
, Garcia-Perez, A.%
\BCBL {}\ \BBA {} Amezquita-Sanchez, J\BPBI P.%
\end{APACrefauthors}%
\unskip\
\newblock
\APACrefYearMonthDay{2025}{}{}.
\newblock
{\BBOQ}\APACrefatitle {Artificial Intelligence-Based Epileptic Seizure Prediction Strategies: A Review} {Artificial intelligence-based epileptic seizure prediction strategies: A review}.{\BBCQ}
\newblock
\APACjournalVolNumPages{AI}{6}{10}{274}.
\PrintBackRefs{\CurrentBib}

\bibitem [\protect \citeauthoryear {%
Popescu%
, Balas%
, Perescu-Popescu%
\BCBL {}\ \BBA {} Mastorakis%
}{%
Popescu%
\ \protect \BOthers {.}}{%
{\protect \APACyear {2009}}%
}]{%
popescu2009multilayer}
\APACinsertmetastar {%
popescu2009multilayer}%
\begin{APACrefauthors}%
Popescu, M\BHBI C.%
, Balas, V\BPBI E.%
, Perescu-Popescu, L.%
\BCBL {}\ \BBA {} Mastorakis, N.%
\end{APACrefauthors}%
\unskip\
\newblock
\APACrefYearMonthDay{2009}{}{}.
\newblock
{\BBOQ}\APACrefatitle {Multilayer perceptron and neural networks} {Multilayer perceptron and neural networks}.{\BBCQ}
\newblock
\APACjournalVolNumPages{WSEAS Transactions on Circuits and Systems}{8}{7}{579--588}.
\PrintBackRefs{\CurrentBib}

\bibitem [\protect \citeauthoryear {%
Radford%
\ \protect \BOthers {.}}{%
Radford%
\ \protect \BOthers {.}}{%
{\protect \APACyear {2019}}%
}]{%
radford2019language}
\APACinsertmetastar {%
radford2019language}%
\begin{APACrefauthors}%
Radford, A.%
, Wu, J.%
, Child, R.%
, Luan, D.%
, Amodei, D.%
, Sutskever, I.%
\BCBL {}\ \BOthersPeriod {.}\end{APACrefauthors}%
\unskip\
\newblock
\APACrefYearMonthDay{2019}{}{}.
\newblock
{\BBOQ}\APACrefatitle {Language models are unsupervised multitask learners} {Language models are unsupervised multitask learners}.{\BBCQ}
\newblock
\APACjournalVolNumPages{OpenAI blog}{1}{8}{9}.
\PrintBackRefs{\CurrentBib}

\bibitem [\protect \citeauthoryear {%
Raymont%
\ \protect \BOthers {.}}{%
Raymont%
\ \protect \BOthers {.}}{%
{\protect \APACyear {2010}}%
}]{%
Raymont2010-ag}
\APACinsertmetastar {%
Raymont2010-ag}%
\begin{APACrefauthors}%
Raymont, V.%
, Salazar, A\BPBI M.%
, Lipsky, R.%
, Goldman, D.%
, Tasick, G.%
\BCBL {}\ \BBA {} Grafman, J.%
\end{APACrefauthors}%
\unskip\
\newblock
\APACrefYearMonthDay{2010}{{\APACmonth{07}}}{}.
\newblock
{\BBOQ}\APACrefatitle {Correlates of posttraumatic epilepsy 35 years following combat brain injury} {Correlates of posttraumatic epilepsy 35 years following combat brain injury}.{\BBCQ}
\newblock
\APACjournalVolNumPages{Neurology}{75}{3}{224--229}.
\PrintBackRefs{\CurrentBib}

\bibitem [\protect \citeauthoryear {%
Reimers%
\ \BBA {} Gurevych%
}{%
Reimers%
\ \BBA {} Gurevych%
}{%
{\protect \APACyear {2019}}%
}]{%
reimers2019sentence}
\APACinsertmetastar {%
reimers2019sentence}%
\begin{APACrefauthors}%
Reimers, N.%
\BCBT {}\ \BBA {} Gurevych, I.%
\end{APACrefauthors}%
\unskip\
\newblock
\APACrefYearMonthDay{2019}{}{}.
\newblock
{\BBOQ}\APACrefatitle {Sentence-bert: Sentence embeddings using siamese bert-networks} {Sentence-bert: Sentence embeddings using siamese bert-networks}.{\BBCQ}
\newblock
\APACjournalVolNumPages{arXiv preprint arXiv:1908.10084}{}{}{}.
\PrintBackRefs{\CurrentBib}

\bibitem [\protect \citeauthoryear {%
Rocca%
\ \protect \BOthers {.}}{%
Rocca%
\ \protect \BOthers {.}}{%
{\protect \APACyear {2019}}%
}]{%
la2019machine}
\APACinsertmetastar {%
la2019machine}%
\begin{APACrefauthors}%
Rocca, M\BPBI L.%
, Garner, R.%
, Jann, K.%
, Kim, H.%
, Vespa, P.%
, Toga, A\BPBI W.%
\BCBL {}\ \BBA {} Duncan, D.%
\end{APACrefauthors}%
\unskip\
\newblock
\APACrefYearMonthDay{2019}{}{}.
\newblock
\APACrefbtitle {Machine learning of multimodal {MRI} to predict the development of epileptic seizures after traumatic brain injury.} {Machine learning of multimodal {MRI} to predict the development of epileptic seizures after traumatic brain injury.}
\newblock
\begin{APACrefURL} \url{https://openreview.net/forum?id=Bye0tkLNcV} \end{APACrefURL}
\PrintBackRefs{\CurrentBib}

\bibitem [\protect \citeauthoryear {%
Saab%
\ \protect \BOthers {.}}{%
Saab%
\ \protect \BOthers {.}}{%
{\protect \APACyear {2024}}%
}]{%
saab2024capabilities}
\APACinsertmetastar {%
saab2024capabilities}%
\begin{APACrefauthors}%
Saab, K.%
, Tu, T.%
, Weng, W\BHBI H.%
, Tanno, R.%
, Stutz, D.%
, Wulczyn, E.%
\BDBL {}others%
\end{APACrefauthors}%
\unskip\
\newblock
\APACrefYearMonthDay{2024}{}{}.
\newblock
{\BBOQ}\APACrefatitle {Capabilities of gemini models in medicine} {Capabilities of gemini models in medicine}.{\BBCQ}
\newblock
\APACjournalVolNumPages{arXiv preprint arXiv:2404.18416}{}{}{}.
\PrintBackRefs{\CurrentBib}

\bibitem [\protect \citeauthoryear {%
Schechter~Vera%
\ \protect \BOthers {.}}{%
Schechter~Vera%
\ \protect \BOthers {.}}{%
{\protect \APACyear {2025}}%
}]{%
embedding_gemma_2025}
\APACinsertmetastar {%
embedding_gemma_2025}%
\begin{APACrefauthors}%
Schechter~Vera, H.%
, Dua, S.%
, Zhang, B.%
, Salz, D.%
, Mullins, R.%
, Raghuram~Panyam, S.%
\BDBL {}Seyedhosseini, M.%
\end{APACrefauthors}%
\unskip\
\newblock
\APACrefYearMonthDay{2025}{}{}.
\newblock
{\BBOQ}\APACrefatitle {EmbeddingGemma: Powerful and Lightweight Text Representations} {Embeddinggemma: Powerful and lightweight text representations}.{\BBCQ}
\newblock
\APACjournalVolNumPages{arXiv preprint arXiv:2509.20354}{}{}{}.
\newblock
\begin{APACrefURL} \url{https://arxiv.org/abs/2509.20354} \end{APACrefURL}
\PrintBackRefs{\CurrentBib}

\bibitem [\protect \citeauthoryear {%
Sollee%
\ \protect \BOthers {.}}{%
Sollee%
\ \protect \BOthers {.}}{%
{\protect \APACyear {2022}}%
}]{%
sollee2022artificial}
\APACinsertmetastar {%
sollee2022artificial}%
\begin{APACrefauthors}%
Sollee, J.%
, Tang, L.%
, Igiraneza, A\BPBI B.%
, Xiao, B.%
, Bai, H\BPBI X.%
\BCBL {}\ \BBA {} Yang, L.%
\end{APACrefauthors}%
\unskip\
\newblock
\APACrefYearMonthDay{2022}{}{}.
\newblock
{\BBOQ}\APACrefatitle {Artificial intelligence for medical image analysis in epilepsy} {Artificial intelligence for medical image analysis in epilepsy}.{\BBCQ}
\newblock
\APACjournalVolNumPages{Epilepsy Research}{}{}{106861}.
\PrintBackRefs{\CurrentBib}

\bibitem [\protect \citeauthoryear {%
Song%
, Zheng%
, Liu%
\BCBL {}\ \BBA {} Gao%
}{%
Song%
\ \protect \BOthers {.}}{%
{\protect \APACyear {2022}}%
}]{%
song2022eeg}
\APACinsertmetastar {%
song2022eeg}%
\begin{APACrefauthors}%
Song, Y.%
, Zheng, Q.%
, Liu, B.%
\BCBL {}\ \BBA {} Gao, X.%
\end{APACrefauthors}%
\unskip\
\newblock
\APACrefYearMonthDay{2022}{}{}.
\newblock
{\BBOQ}\APACrefatitle {EEG conformer: Convolutional transformer for EEG decoding and visualization} {Eeg conformer: Convolutional transformer for eeg decoding and visualization}.{\BBCQ}
\newblock
\APACjournalVolNumPages{IEEE Transactions on Neural Systems and Rehabilitation Engineering}{31}{}{710--719}.
\PrintBackRefs{\CurrentBib}

\bibitem [\protect \citeauthoryear {%
Sounack%
\ \protect \BOthers {.}}{%
Sounack%
\ \protect \BOthers {.}}{%
{\protect \APACyear {2025}}%
}]{%
sounack2025bioclinicalmodernbertstateoftheartlongcontext}
\APACinsertmetastar {%
sounack2025bioclinicalmodernbertstateoftheartlongcontext}%
\begin{APACrefauthors}%
Sounack, T.%
, Davis, J.%
, Durieux, B.%
, Chaffin, A.%
, Pollard, T\BPBI J.%
, Lehman, E.%
\BDBL {}Lindvall, C.%
\end{APACrefauthors}%
\unskip\
\newblock
\APACrefYearMonthDay{2025}{}{}.
\newblock
\APACrefbtitle {BioClinical ModernBERT: A State-of-the-Art Long-Context Encoder for Biomedical and Clinical NLP.} {Bioclinical modernbert: A state-of-the-art long-context encoder for biomedical and clinical nlp.}
\newblock
\begin{APACrefURL} \url{https://arxiv.org/abs/2506.10896} \end{APACrefURL}
\PrintBackRefs{\CurrentBib}

\bibitem [\protect \citeauthoryear {%
Taspinar%
\ \BBA {} Ozkurt%
}{%
Taspinar%
\ \BBA {} Ozkurt%
}{%
{\protect \APACyear {2024}}%
}]{%
taspinar2024review}
\APACinsertmetastar {%
taspinar2024review}%
\begin{APACrefauthors}%
Taspinar, G.%
\BCBT {}\ \BBA {} Ozkurt, N.%
\end{APACrefauthors}%
\unskip\
\newblock
\APACrefYearMonthDay{2024}{}{}.
\newblock
{\BBOQ}\APACrefatitle {A review of ADHD detection studies with machine learning methods using rsfMRI data} {A review of adhd detection studies with machine learning methods using rsfmri data}.{\BBCQ}
\newblock
\APACjournalVolNumPages{NMR in Biomedicine}{37}{8}{e5138}.
\PrintBackRefs{\CurrentBib}

\bibitem [\protect \citeauthoryear {%
Thawani%
, Pujara%
, Szekely%
\BCBL {}\ \BBA {} Ilievski%
}{%
Thawani%
\ \protect \BOthers {.}}{%
{\protect \APACyear {2021}}%
}]{%
thawani2021representingnumbersnlpsurvey}
\APACinsertmetastar {%
thawani2021representingnumbersnlpsurvey}%
\begin{APACrefauthors}%
Thawani, A.%
, Pujara, J.%
, Szekely, P\BPBI A.%
\BCBL {}\ \BBA {} Ilievski, F.%
\end{APACrefauthors}%
\unskip\
\newblock
\APACrefYearMonthDay{2021}{}{}.
\newblock
\APACrefbtitle {Representing Numbers in NLP: a Survey and a Vision.} {Representing numbers in nlp: a survey and a vision.}
\newblock
\begin{APACrefURL} \url{https://arxiv.org/abs/2103.13136} \end{APACrefURL}
\PrintBackRefs{\CurrentBib}

\bibitem [\protect \citeauthoryear {%
Thirunavukarasu%
\ \protect \BOthers {.}}{%
Thirunavukarasu%
\ \protect \BOthers {.}}{%
{\protect \APACyear {2023}}%
}]{%
thirunavukarasu2023large}
\APACinsertmetastar {%
thirunavukarasu2023large}%
\begin{APACrefauthors}%
Thirunavukarasu, A\BPBI J.%
, Ting, D\BPBI S\BPBI J.%
, Elangovan, K.%
, Gutierrez, L.%
, Tan, T\BPBI F.%
\BCBL {}\ \BBA {} Ting, D\BPBI S\BPBI W.%
\end{APACrefauthors}%
\unskip\
\newblock
\APACrefYearMonthDay{2023}{}{}.
\newblock
{\BBOQ}\APACrefatitle {Large language models in medicine} {Large language models in medicine}.{\BBCQ}
\newblock
\APACjournalVolNumPages{Nature medicine}{29}{8}{1930--1940}.
\PrintBackRefs{\CurrentBib}

\bibitem [\protect \citeauthoryear {%
Vasantharajan%
\ \protect \BOthers {.}}{%
Vasantharajan%
\ \protect \BOthers {.}}{%
{\protect \APACyear {2022}}%
}]{%
9980157}
\APACinsertmetastar {%
9980157}%
\begin{APACrefauthors}%
Vasantharajan, C.%
, Tun, K\BPBI Z.%
, Thi-Nga, H.%
, Jain, S.%
, Rong, T.%
\BCBL {}\ \BBA {} Siong, C\BPBI E.%
\end{APACrefauthors}%
\unskip\
\newblock
\APACrefYearMonthDay{2022}{}{}.
\newblock
{\BBOQ}\APACrefatitle {MedBERT: A Pre-trained Language Model for Biomedical Named Entity Recognition} {Medbert: A pre-trained language model for biomedical named entity recognition}.{\BBCQ}
\newblock
\BIn{} \APACrefbtitle {2022 Asia-Pacific Signal and Information Processing Association Annual Summit and Conference (APSIPA ASC)} {2022 asia-pacific signal and information processing association annual summit and conference (apsipa asc)}\ (\BPG~1482-1488).
\newblock
\begin{APACrefDOI} 10.23919/APSIPAASC55919.2022.9980157 \end{APACrefDOI}
\PrintBackRefs{\CurrentBib}

\bibitem [\protect \citeauthoryear {%
Vaswani%
\ \protect \BOthers {.}}{%
Vaswani%
\ \protect \BOthers {.}}{%
{\protect \APACyear {2017}}%
}]{%
vaswani2017attention}
\APACinsertmetastar {%
vaswani2017attention}%
\begin{APACrefauthors}%
Vaswani, A.%
, Shazeer, N.%
, Parmar, N.%
, Uszkoreit, J.%
, Jones, L.%
, Gomez, A\BPBI N.%
\BDBL {}Polosukhin, I.%
\end{APACrefauthors}%
\unskip\
\newblock
\APACrefYearMonthDay{2017}{}{}.
\newblock
{\BBOQ}\APACrefatitle {Attention is all you need} {Attention is all you need}.{\BBCQ}
\newblock
\APACjournalVolNumPages{Advances in neural information processing systems}{30}{}{}.
\PrintBackRefs{\CurrentBib}

\bibitem [\protect \citeauthoryear {%
Verellen%
\ \BBA {} Cavazos%
}{%
Verellen%
\ \BBA {} Cavazos%
}{%
{\protect \APACyear {2010}}%
}]{%
verellen2010post}
\APACinsertmetastar {%
verellen2010post}%
\begin{APACrefauthors}%
Verellen, R\BPBI M.%
\BCBT {}\ \BBA {} Cavazos, J\BPBI E.%
\end{APACrefauthors}%
\unskip\
\newblock
\APACrefYearMonthDay{2010}{}{}.
\newblock
{\BBOQ}\APACrefatitle {Post-traumatic epilepsy: an overview} {Post-traumatic epilepsy: an overview}.{\BBCQ}
\newblock
\APACjournalVolNumPages{Therapy}{7}{5}{527}.
\PrintBackRefs{\CurrentBib}

\bibitem [\protect \citeauthoryear {%
X.~Wang%
, Zhong%
, Lei%
, Chen%
\BCBL {}\ \protect \BOthers {.}}{%
X.~Wang%
, Zhong%
, Lei%
, Chen%
\BCBL {}\ \protect \BOthers {.}}{%
{\protect \APACyear {2021}}%
}]{%
wang2021artificial}
\APACinsertmetastar {%
wang2021artificial}%
\begin{APACrefauthors}%
Wang, X.%
, Zhong, J.%
, Lei, T.%
, Chen, D.%
, Wang, H.%
, Zhu, L.%
\BDBL {}Liu, L.%
\end{APACrefauthors}%
\unskip\
\newblock
\APACrefYearMonthDay{2021}{}{}.
\newblock
{\BBOQ}\APACrefatitle {An artificial neural network prediction model for posttraumatic epilepsy: retrospective cohort study} {An artificial neural network prediction model for posttraumatic epilepsy: retrospective cohort study}.{\BBCQ}
\newblock
\APACjournalVolNumPages{Journal of Medical Internet Research}{23}{8}{e25090}.
\PrintBackRefs{\CurrentBib}

\bibitem [\protect \citeauthoryear {%
X.~Wang%
, Zhong%
, Lei%
, Wang%
\BCBL {}\ \protect \BOthers {.}}{%
X.~Wang%
, Zhong%
, Lei%
, Wang%
\BCBL {}\ \protect \BOthers {.}}{%
{\protect \APACyear {2021}}%
}]{%
wang2021development}
\APACinsertmetastar {%
wang2021development}%
\begin{APACrefauthors}%
Wang, X.%
, Zhong, J.%
, Lei, T.%
, Wang, H\BHBI j.%
, Zhu, L\BHBI n.%
, Chu, S.%
\BDBL {}Liu, L.%
\end{APACrefauthors}%
\unskip\
\newblock
\APACrefYearMonthDay{2021}{}{}.
\newblock
{\BBOQ}\APACrefatitle {Development and external validation of a predictive nomogram model of posttraumatic epilepsy: a retrospective analysis} {Development and external validation of a predictive nomogram model of posttraumatic epilepsy: a retrospective analysis}.{\BBCQ}
\newblock
\APACjournalVolNumPages{Seizure}{88}{}{36--44}.
\PrintBackRefs{\CurrentBib}

\bibitem [\protect \citeauthoryear {%
Z.~Wang%
, Gao%
, Xiao%
\BCBL {}\ \BBA {} Sun%
}{%
Z.~Wang%
\ \protect \BOthers {.}}{%
{\protect \APACyear {2024}}%
}]{%
wang2024meditabscalingmedicaltabular}
\APACinsertmetastar {%
wang2024meditabscalingmedicaltabular}%
\begin{APACrefauthors}%
Wang, Z.%
, Gao, C.%
, Xiao, C.%
\BCBL {}\ \BBA {} Sun, J.%
\end{APACrefauthors}%
\unskip\
\newblock
\APACrefYearMonthDay{2024}{}{}.
\newblock
\APACrefbtitle {MediTab: Scaling Medical Tabular Data Predictors via Data Consolidation, Enrichment, and Refinement.} {Meditab: Scaling medical tabular data predictors via data consolidation, enrichment, and refinement.}
\newblock
\begin{APACrefURL} \url{https://arxiv.org/abs/2305.12081} \end{APACrefURL}
\PrintBackRefs{\CurrentBib}

\bibitem [\protect \citeauthoryear {%
Yang%
\ \protect \BOthers {.}}{%
Yang%
\ \protect \BOthers {.}}{%
{\protect \APACyear {2022}}%
}]{%
yang2022large}
\APACinsertmetastar {%
yang2022large}%
\begin{APACrefauthors}%
Yang, X.%
, Chen, A.%
, PourNejatian, N.%
, Shin, H\BPBI C.%
, Smith, K\BPBI E.%
, Parisien, C.%
\BDBL {}others%
\end{APACrefauthors}%
\unskip\
\newblock
\APACrefYearMonthDay{2022}{}{}.
\newblock
{\BBOQ}\APACrefatitle {A large language model for electronic health records} {A large language model for electronic health records}.{\BBCQ}
\newblock
\APACjournalVolNumPages{npj Digital Medicine}{5}{1}{194}.
\newblock
\begin{APACrefDOI} 10.1038/s41746-022-00742-2 \end{APACrefDOI}
\PrintBackRefs{\CurrentBib}

\bibitem [\protect \citeauthoryear {%
Zhang%
\ \protect \BOthers {.}}{%
Zhang%
\ \protect \BOthers {.}}{%
{\protect \APACyear {2025}}%
}]{%
qwen3embedding}
\APACinsertmetastar {%
qwen3embedding}%
\begin{APACrefauthors}%
Zhang, Y.%
, Li, M.%
, Long, D.%
, Zhang, X.%
, Lin, H.%
, Yang, B.%
\BDBL {}Zhou, J.%
\end{APACrefauthors}%
\unskip\
\newblock
\APACrefYearMonthDay{2025}{}{}.
\newblock
{\BBOQ}\APACrefatitle {Qwen3 Embedding: Advancing Text Embedding and Reranking Through Foundation Models} {Qwen3 embedding: Advancing text embedding and reranking through foundation models}.{\BBCQ}
\newblock
\APACjournalVolNumPages{arXiv preprint arXiv:2506.05176}{}{}{}.
\PrintBackRefs{\CurrentBib}

\bibitem [\protect \citeauthoryear {%
B.~Zhou%
\ \protect \BOthers {.}}{%
B.~Zhou%
\ \protect \BOthers {.}}{%
{\protect \APACyear {2020}}%
}]{%
zhou2020machine}
\APACinsertmetastar {%
zhou2020machine}%
\begin{APACrefauthors}%
Zhou, B.%
, An, D.%
, Xiao, F.%
, Niu, R.%
, Li, W.%
, Li, W.%
\BDBL {}others%
\end{APACrefauthors}%
\unskip\
\newblock
\APACrefYearMonthDay{2020}{}{}.
\newblock
{\BBOQ}\APACrefatitle {Machine learning for detecting mesial temporal lobe epilepsy by structural and functional neuroimaging} {Machine learning for detecting mesial temporal lobe epilepsy by structural and functional neuroimaging}.{\BBCQ}
\newblock
\APACjournalVolNumPages{Frontiers of Medicine}{14}{5}{630--641}.
\PrintBackRefs{\CurrentBib}

\bibitem [\protect \citeauthoryear {%
M.~Zhou%
\ \protect \BOthers {.}}{%
M.~Zhou%
\ \protect \BOthers {.}}{%
{\protect \APACyear {2018}}%
}]{%
zhou2018epileptic}
\APACinsertmetastar {%
zhou2018epileptic}%
\begin{APACrefauthors}%
Zhou, M.%
, Tian, C.%
, Cao, R.%
, Wang, B.%
, Niu, Y.%
, Hu, T.%
\BDBL {}Xiang, J.%
\end{APACrefauthors}%
\unskip\
\newblock
\APACrefYearMonthDay{2018}{}{}.
\newblock
{\BBOQ}\APACrefatitle {Epileptic seizure detection based on EEG signals and CNN} {Epileptic seizure detection based on eeg signals and cnn}.{\BBCQ}
\newblock
\APACjournalVolNumPages{Frontiers in neuroinformatics}{12}{}{95}.
\PrintBackRefs{\CurrentBib}

\end{thebibliography}

\appendix
\section{Additional Details}
\subsection{Architectural and Pre-training Details of LLMs Used in Our Experiments}
A transformer is a type of neural network architecture designed to model sequences of information, such as text or timeseries, by learning relationships between all parts of the input at once. At the core of a transformer are attention layers, which allow the model to focus on the most relevant parts of the input when forming predictions. Within each attention layer, the model uses multiple attention heads, where each head learns to capture a different aspect of information in the data. This multi-head design enables the model to investigate the same data from several complementary perspectives simultaneously. 
Information processed by the transformer is represented as numerical vectors known as embeddings, and the size of these vectors is referred to as the embedding dimension. A larger embedding dimension allows the model to encode richer and more nuanced information about each clinical variable. Together, attention layers, multiple attention heads, and sufficiently expressive embeddings enable transformers to capture complex and long-range patterns in clinical data that are difficult to model with traditional approaches.

MedBERT \cite{9980157} \footnote{https://huggingface.co/Charangan/MedBERT} is a transformer-based encoder model specifically optimized for biomedical named entity recognition (NER). Unlike other versions of MedBERT designed for structured record analysis, model used in this study follows the standard BERT-base architecture (12 layers, 768 hidden dimensions) but is initialized with weights from Bio\_ClinicalBERT to leverage existing clinical knowledge. It was further pre-trained on a consolidated biomedical corpus totaling 57.46 million tokens, sourced from the National NLP Clinical Challenges (N2C2), the BioNLP project, and the CRAFT corpus. To enhance its understanding of general medical concepts, the pre-training data was augmented with 10,000 medical-related Wikipedia pages, filtered for domain relevance using the XenC cross-entropy tool. The model was trained for 200,000 steps with a masked language modeling (MLM) objective, masking probability of 0.15, and a maximum sequence length of 256 tokens.

PubMedBERT \cite{mezzetti2023embeddings} \footnote{https://huggingface.co/NeuML/pubmedbert-base-embeddings} follows the standard BERT-base \cite{devlin2019bertpretrainingdeepbidirectional} configuration, consisting of 12 Transformer layers, a hidden dimension of 768, and 12 attention heads. Unlike models that employ continual pre-training from general-domain weights, PubMedBERT was trained from scratch specifically on the biomedical domain. This allowed for the generation of a specialized vocabulary of 30,000 tokens derived directly from 14 million PubMed abstracts and over 3 million PMC full-text articles. The model was optimized using a standard Masked Language Modeling (MLM) objective on a total of approximately 14 billion tokens, ensuring that its sub-word tokenization and contextual representations are highly sensitive to technical medical terminology.

MedCPT-Article-Encoder \cite{jin2023medcpt} \footnote{https://huggingface.co/ncbi/MedCPT-Article-Encoder} is a specialized discriminative model designed for dense biomedical information retrieval. Architecturally, it is initialized from the PubMedBERT checkpoint and functions as part of a bi-encoder framework. The model maps article components (titles and abstracts) into a fixed 768-dimensional vector space. Its pre-training uses contrastive learning applied to a massive corpus of 255 million user search logs from PubMed. By optimizing a contrastive loss that maximizes the cosine similarity between search queries and clicked articles while minimizing it for in-batch negatives, the encoder learns to prioritize semantic relevance and clinical retrieval over generic linguistic similarity.

BioClinical-ModernBERT \cite{sounack2025bioclinicalmodernbertstateoftheartlongcontext} \footnote{https://huggingface.co/thomas-sounack/BioClinical-ModernBERT-base} utilizes the more advanced ModernBERT architecture, which introduces several optimizations for long-context efficiency. It features an expanded sequence length of 8,192 tokens, facilitated by Rotary Positional Embeddings (RoPE) and a hybrid attention mechanism that alternates between local (unpadding) and global attention. Pre-training involved a two-phase regime on 53.5 billion tokens. The initial phase focused on a broad mixture of PubMed and PMC data, while the second phase utilized 20 distinct clinical datasets (totaling 2.8 billion tokens) for domain refinement. Notably, the model employed a 30\% masking ratio during MLM, which is double the standard BERT rate, to force the encoder to learn deeper structural dependencies within complex clinical narratives.

EmbeddingGemma-300M \cite{embedding_gemma_2025} is a 24-layer encoder-only model derived from the Gemma 3 family. It generates embeddings with a dimension of 768 and supports a context window of 8,192 tokens. The model employs a two-stage linear projection and was pre-trained on 320 billion tokens using a hybrid objective of Noise-Contrastive Estimation and Geometric Embedding Distillation from larger teacher models.
Google text-embedding-004 \cite{google_gemini_embeddings_2024} is a bidirectional Transformer model initialized from model weights of the Gemini family. It supports a large context window of 20,480 tokens. It uses Matryoshka Representation Learning \cite{kusupati2024matryoshkarepresentationlearning}, which allows the embeddings to be truncated to lower dimensions, such as 768, while maintaining high retrieval performance across multilingual benchmarks. 
BGE-M3 \cite{bge-m3} is a transformer encoder model based on the XLM-RoBERTa backbone, producing an embedding dimension of 1,024 and a context window of 8,192 tokens. It is optimized for multi-functionality via a joint training objective that combines dense retrieval, lexical retrieval, and multi-vector late-interaction retrieval. It was trained on 1.2 billion multilingual text pairs using a self-knowledge distillation approach. 
Qwen-0.6B/4B/8B \cite{qwen3embedding} is a dense encoder model built on the Qwen3 architecture, featuring 28 layers and an embedding dimension of 1,024/2560/4096. It is designed for long-document processing with a context window of 32,768 tokens. The model is instruction-aware, allowing the embedding geometry to be guided by task-specific prefixes, and was trained using a multi-stage contrastive pipeline followed by instruction fine-tuning. 
LLaMA-8B \cite{babakhin2025llamaembednemotron8buniversaltextembedding} (Llama-Embed-Nemotron-8B) is a universal embedding model derived from the Llama-3.1-8B decoder model. The architecture was converted to a bidirectional encoder by removing the causal mask, resulting in 32 layers and an embedding dimension of 4,096. It utilizes a context window of 8,192 tokens and was trained via contrastive learning on a mixture of 16.1 million query-document pairs, including extensive synthetic data to ensure robust cross-lingual retrieval.

\subsection{Implementation Details of Classifiers}
All classifiers were evaluated using stratified 5-fold cross-validation, repeated across 30 random seeds to ensure robustness and statistical stability. 
To handle the high dimensionality of embedding features, a per-aspect PCA was performed within each training fold, reducing each feature set to 16 components prior to concatenation. 
The XGBoost classifier was configured with a learning rate of 0.05 and a maximum tree depth of 3, and employed both L1 and L2 regularization (0.5 and 1.0, respectively) to mitigate overfitting. A total of 2,000 boosting rounds were used, with early stopping applied after 50 rounds based on validation performance. 
The MLP was implemented in PyTorch and consisted of two fully connected hidden layers with 256 and 128 hidden dimensions, respectively, followed by batch normalization and ReLU activation. A dropout rate of 0.3 was applied to reduce overfitting. The network was trained for 50 epochs using the Adam optimizer with a learning rate of 0.001 and a batch size of 64. 
The SVM employed a Radial Basis Function (RBF) kernel with the regularization parameter $C=1.0$ and a scaled kernel coefficient. To address class imbalance, all models incorporated class-aware weighting strategies, with loss functions or decision boundaries adjusted using weights computed from the ratio of negative to positive samples within each training fold.

\end{document}